\pdfoutput=1
\documentclass[11pt]{article}

\usepackage{EMNLP2022}

\usepackage{times}
\usepackage{latexsym}
\usepackage{graphicx}
\usepackage{float}
\usepackage{subfigure}
\usepackage{multirow}
\usepackage{booktabs}

\usepackage{soul}
\usepackage{multirow}
\usepackage{algorithm}
\usepackage{algpseudocode}
\usepackage{amsmath}

\usepackage{colortbl} 
\usepackage{xcolor}
\usepackage{array}   
\usepackage{enumitem}
\usepackage[inkscapeformat=png]{svg}

\usepackage[T1]{fontenc}
\usepackage[utf8]{inputenc}

\usepackage{microtype}

\usepackage{inconsolata}

\newcommand{\wan}[1]{\textcolor{black}{#1}}

\title{G-MAP: General Memory-Augmented Pre-trained Language Model \\ for Domain Tasks}

\author{Zhongwei Wan\textsuperscript{1,2}\thanks{$^*$ This work is done when Zhongwei Wan is an intern at Huawei Noah’s Ark Lab},  Yichun Yin\textsuperscript{3}, Wei Zhang\textsuperscript{3}, Jiaxin Shi\textsuperscript{4}, Lifeng Shang\textsuperscript{3},\\ 
{\bf Guangyong Chen}\textsuperscript{5}\thanks{$\dag$ Guangyong Chen is the corresponding author}, 
{\bf Xin Jiang}\textsuperscript{3}, {\bf Qun Liu}\textsuperscript{3} \\
\textsuperscript{1} Guangdong Provincial Key Laboratory of Computer Vision and Virtual Reality
Technology, \\ Shenzhen Institute of Advanced Technology, 
Chinese Academy of Science \\
\textsuperscript{2}University of Chinese Academy of Sciences 
\textsuperscript{3}Huawei Noah's Ark Lab\\
\textsuperscript{4}Huawei Cloud Computing 
\textsuperscript{5}Zhejiang Lab\\
zw.wan1@siat.ac.cn, gychen@zhejianglab.com \\
\{yinyichun, zhangwei379, shijiaxin3, shang.lifeng, jiang.xin, qun.liu\}@huawei.com
}

\begin{document}

\maketitle

\begin{abstract}

Recently, domain-specific PLMs have been proposed to boost the task performance of specific domains (e.g., biomedical and computer science) by continuing to pre-train general PLMs with domain-specific corpora. However, this Domain-Adaptive Pre-Training (DAPT; \citet{DBLP:conf/acl/GururanganMSLBD20}) tends to forget the previous general knowledge acquired by general PLMs, which leads to a \emph{catastrophic forgetting} phenomenon and sub-optimal performance. To alleviate this problem, we propose a new framework of \textbf{G}eneral \textbf{M}emory-\textbf{A}ugmented \textbf{P}re-trained Language Model (\textbf{G-MAP}), which augments the domain-specific PLM by a memory representation built from the frozen general PLM without losing any general knowledge. Specifically, we propose a new memory-augmented layer, and based on it, different augmented strategies are explored to build the memory representation and then adaptively fuse it into the domain-specific PLM. We demonstrate the effectiveness of G-MAP on various domains (biomedical and computer science publications, news, and reviews) and different kinds (text classification, QA, NER) of tasks, and the extensive results show that the proposed G-MAP\footnote{https://github.com/SUSTechBruce/G-MAP.} can achieve SOTA results on all tasks. 

\end{abstract}

\section{Introduction}
Pre-trained Language models (PLMs), such as BERT~\cite{DBLP:conf/naacl/DevlinCLT19} and RoBERTa~\cite{DBLP:journals/corr/abs-1907-11692}, have achieved promising performance on NLP tasks. Typically, these general models are firstly pre-trained on large unlabeled corpus and then directly fine-tuned on downstream tasks. However, there is an inherent gap in text distribution between unlabeled pre-training corpus and labeled task corpus, which leads to the distribution shift problem ~\cite{DBLP:conf/acl/GururanganMSLBD20} and makes PLMs perform poorly on some domain tasks~\cite{DBLP:conf/emnlp/BeltagyLC19, DBLP:journals/bioinformatics/LeeYKKKSK20}. To address this shift problem, the domain-adaptive pre-training (DAPT) is proposed~\cite{DBLP:journals/corr/abs-1904-05342,DBLP:conf/emnlp/BeltagyLC19,DBLP:conf/acl/GururanganMSLBD20, DBLP:journals/bioinformatics/LeeYKKKSK20} to further pretrain general PLMs on large-scale domain corpora, achieving better performance than general PLMs.

\begin{figure}[tbp]
\centering
\begin{minipage}[c]{0.16\textwidth}
\centering
\includegraphics[width=2.6cm,height = 2.6cm]{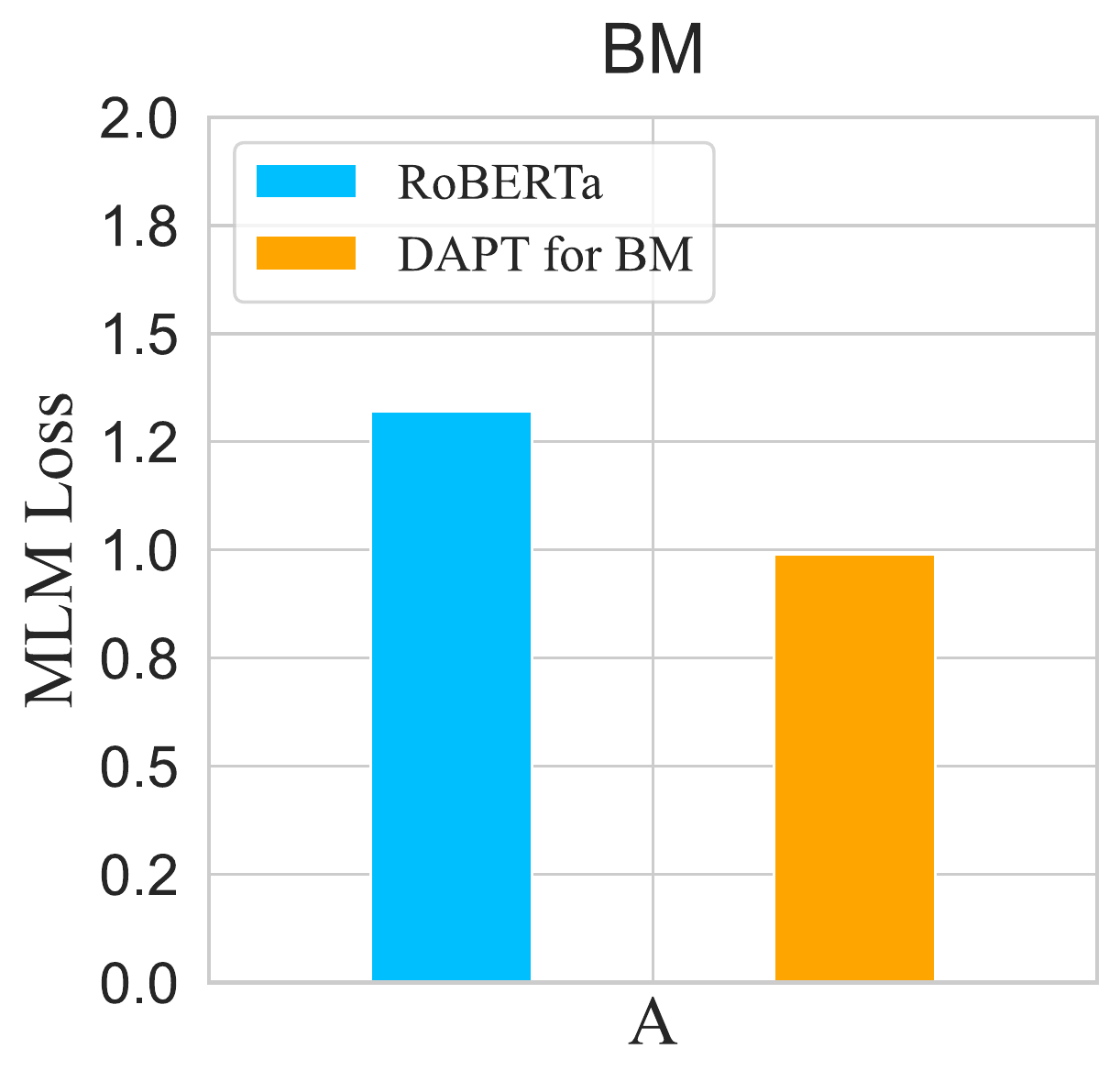}
\end{minipage}
\begin{minipage}[c]{0.15\textwidth}
\centering
\includegraphics[width=2.5cm,height = 2.6cm]{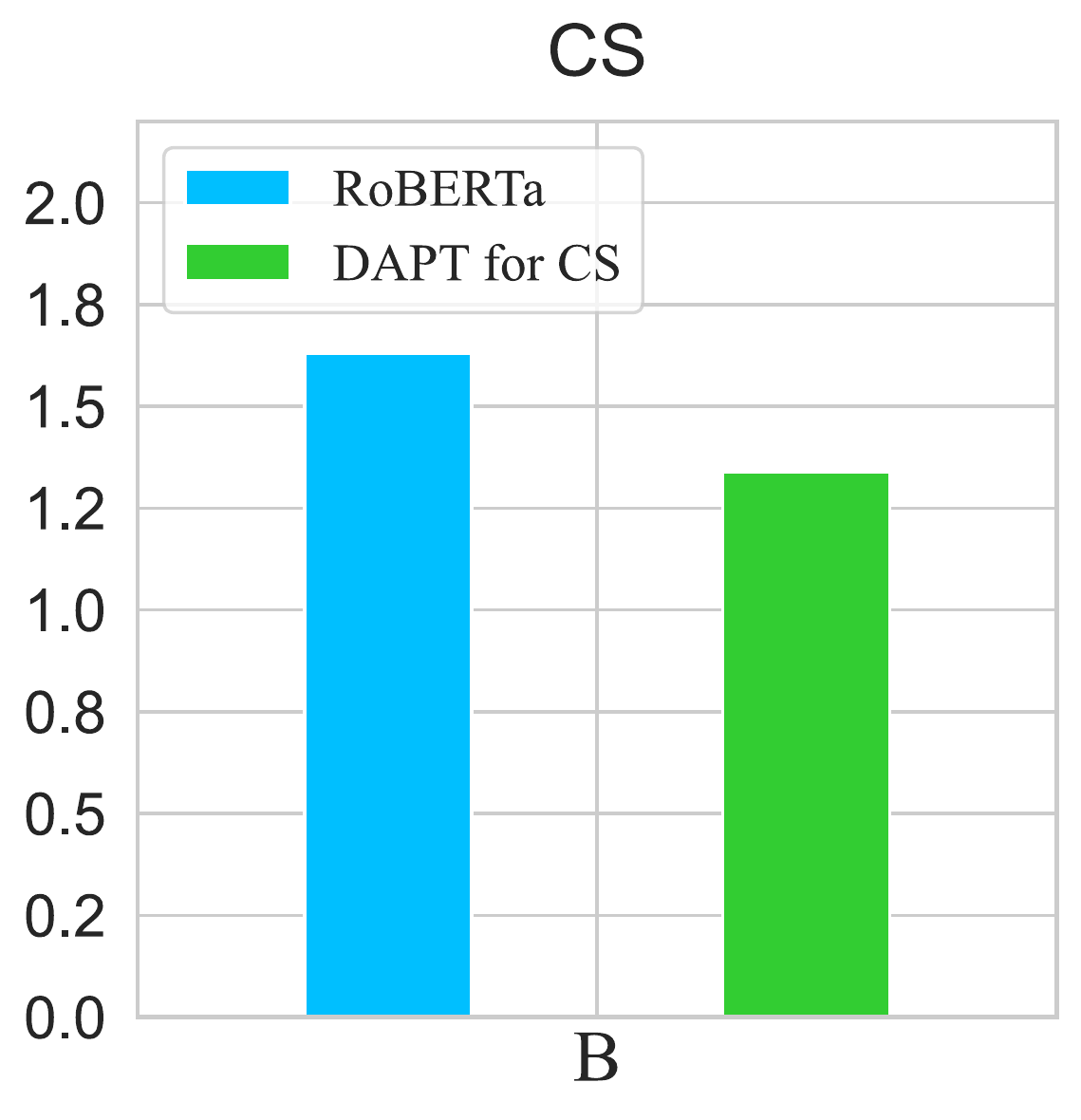}
\end{minipage}
\begin{minipage}[c]{0.15\textwidth}
\centering
\includegraphics[width=2.6cm,height = 2.6cm]{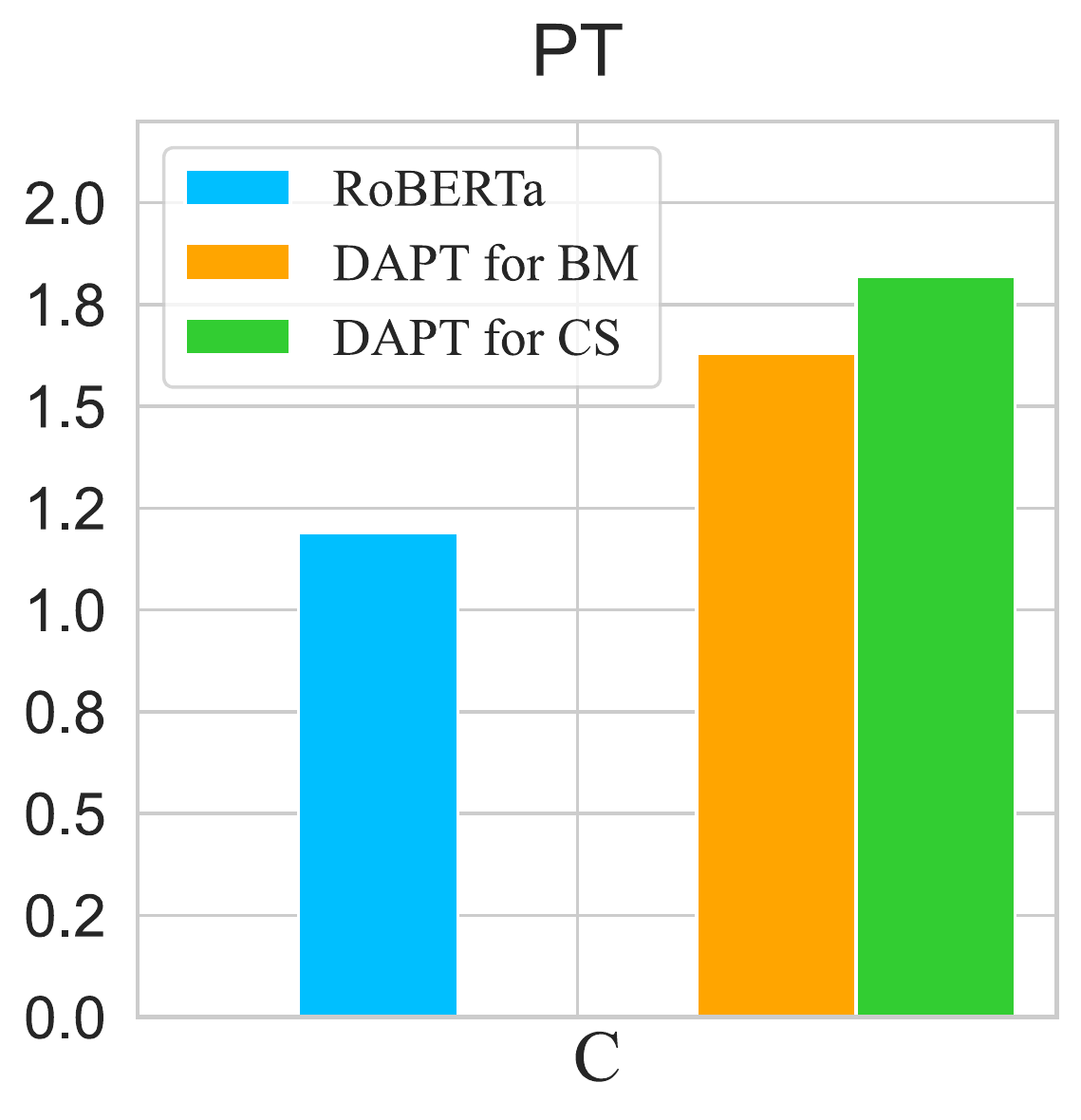}
\end{minipage}
\caption{Masked LM (MLM) loss of RoBERTa on 50K randomly sampled documents from each domain before and after DAPT. Figure A and B denote the inference loss of general RoBERTa-base and domain-specific PLMs on the samples of biomedical (BM) and computer science (CS). Figure C means the loss of these models on the samples from the pre-training (PT) corpus of RoBERTa. We report the results of  \cite{DBLP:conf/acl/GururanganMSLBD20} and lower MLM loss is better.}\label{fig:MLM_loss}
\vspace{-0.2in}
\end{figure}

Although DAPT can effectively learn the domain distribution of the target task, its continual pretraining process updates the parameters of general PLMs, which inevitably leads to partial general knowledge being forgotten. This  \emph{catastrophic forgetting} \cite{DBLP:journals/corr/GoodfellowMDCB13, DBLP:conf/eccv/LiH16, DBLP:conf/naacl/ThompsonGKDK19} phenomenon is verified in Figure~\ref{fig:MLM_loss}, where we observe that the domain-specific PLMs show better results than general PLMs on domain corpus, but perform worse on the general corpus. We argue that this forgotten knowledge is beneficial for domain-specific PLMs and should be used to improve their generalization ability on domain tasks. 

To alleviate the \emph{catastrophic forgetting}, we propose a simple yet effective memory-augmented framework named \textbf{G}eneral \textbf{M}emory-\textbf{A}ugmented \textbf{P}re-trained model (\textbf{G-MAP}). In addition to the backbone domain-specific PLM, G-MAP introduces a new memory-augmented layer. It explicitly incorporates the representation built from a frozen general PLM as the memory to make the backbone model access the complete general knowledge. Then, a new proposed memory-attention within the memory-augmented layer enables the domain-specific PLM adaptively combine the memory representation and the domain-specific representation. Using the memory built from the frozen general PLM has two advantages: (1) frozen PLM never suffers from forgetfulness since the parameters remain unchanged~\cite{Levine}; (2) it doesn't require additional training for the general PLM during fine-tuning. However, building and fusing memory into a backbone model is essentially a many-to-many scenario, where we need to choose which layer output of the general PLM as the memory representation, and which layer in the domain-specific PLM should be fused. Thus, we propose several memory-augmented strategies for better building and then combining the memory representation into domain-specific PLM.

% \gychen{Catastrophic forgetting or continual learning is a widely studied problem in machine learning community, among which memory based methods may be the most fundamental. However, in this paper, we propose some different memory fusion mechanism. I think we should briefly survey some well known continual learning papers here to better emphasize our contribution.}

%  \gychen{I think we should clarify that this idea has not been widely used and systematically investigated in continual learning.}

We evaluate our G-MAP on text classification, Question Answering (QA) and Name Entity Recognition (NER) tasks covering four domains: biomedical science, computer science, news, and reviews. Experimental results demonstrate that G-MAP outperforms existing baselines on all tasks. We compare different memory-augmented strategies, and the results show that the proposed chunk-based gated memory transfer strategy achieves the best results. In addition, for the memory representation building, we empirically find that the freezing way is better than the unfreezing one, which also has better training efficiency. Furthermore, we apply the proposed framework to a small-scale domain pre-training setting and find that G-MAP is also practical in achieving lower MLM loss. Our contributions are summarized below:
\vspace{-1mm}
\begin{itemize}[leftmargin=*]
   \setlength{\itemsep}{1.0pt}
   \setlength{\parsep}{1.0pt}
   \setlength{\parskip}{1.0pt}
   
    \item We empirically find that forgotten general knowledge due to \emph{catastrophic forgetting} can benefit the domain-specific downstream tasks since it can improve PLMs' generalization ability.
    
    \item We propose a novel G-MAP framework, which introduces several memory-augmented strategies to construct the memory representation from the frozen general PLM effectively and then fuse it into the domain-specific PLM by a new memory-augmented layer.
    
    \item We conduct extensive experiments on various domain-specific tasks, including text classification, QA, and NER, the results demonstrating that our G-MAP outperforms existing baselines.
\end{itemize}

% which utilizes \emph{Memory-Attention} to self-adaptively combine generic memory from general frozen model to remedy the forgotten domain knowledge after DAPT as well as  augment generalization ability of domain-specific PLM during fine-tune.

\begin{figure}
  \centering
\includegraphics[width=1\linewidth]{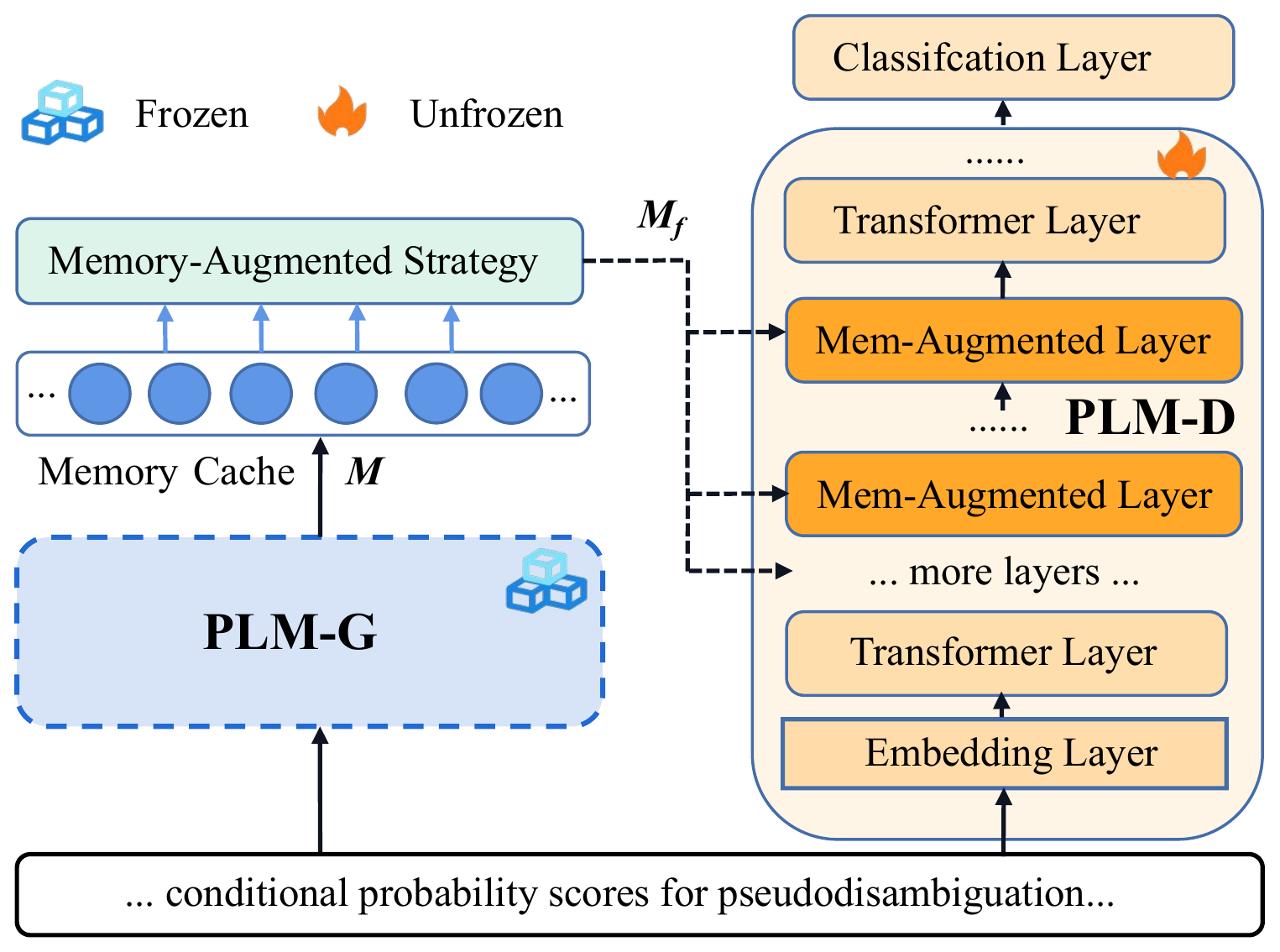}
  \vspace{-0.15in}
  \caption{A framework of G-MAP with the cs-domain task input. PLM-G denotes the frozen general PLM, PLM-D denotes the domain-specific PLM.}
  \label{fig:simple_arch}
  \vspace{-0.15in}
\end{figure}

\section{The Method of G-MAP}
In this section, we first overview the G-MAP framework. Then we detail a new memory-augmented layer that fuses general knowledge into domain-specific PLMs. Finally, we propose different memory-augmented strategies, including single-layer memory transfer, multiple-layer memory transfer, gated memory transfer, and chunk-based gated memory transfer.

\subsection{Overview}
Our G-MAP framework aims to tackle the \emph{catastrophic forgetting} of domain-specific PLMs by using the memory cache built from the frozen general PLMs, which is illustrated in Figure~\ref{fig:simple_arch}. Given an sequence $\boldsymbol{x} = \left[x_1,x_2, \ldots, x_t, \ldots, x_n \right]$ with $\boldsymbol{x_t}$ denoting the \emph{t}-th token, general PLMs output the contextual representations of the input tokens as the memory cache, which is fed into the domain-specific PLMs to build final representation for domain tasks:

\vspace{-3mm}
{\small
\begin{align} 
\boldsymbol{M}&=\operatorname{PLM-G}\left(\boldsymbol{x} ; \theta_{g}\right)\\
\boldsymbol{H}&=\operatorname{PLM-D}\left(\boldsymbol{x}, \boldsymbol{M}; \theta_{d}\right)
\end{align}
}
\nobreak
\vspace{-4mm}

\noindent where $\theta_g$ and $\theta_d$ are the parameters of general and domain-specific PLMs, respectively. We only update the $\theta_d$ and the $\theta_g$ is frozen when fine-tuning. The general PLM could be a BERT or RoBERTa, which contains $l$ layers of Transformer \cite{DBLP:conf/nips/VaswaniSPUJGKP17} encoder blocks and outputs a set of hidden states denoted as \wan{a memory cache} $\boldsymbol{M}=\left\{\boldsymbol{M^{1},M^{2}, \ldots, M^{l}}\right\}$. 
In the G-MAP framework, the domain-specific PLM utilizes a new memory-augmented layer to adaptively incorporate the memory representation  built from the memory cache $\boldsymbol{M}$ and enhance its generalization ability. Note that memory-augmented layer is built by only the parameters of the original Transformer layer without adding new ones. Moreover, we explore different memory-augmented strategies and further propose the chunk-based memory transfer strategy, which fully uses the memories from different chunks of the general PLM.

\begin{figure*}[htbp]
  \centering
  \includegraphics[width=1\linewidth]{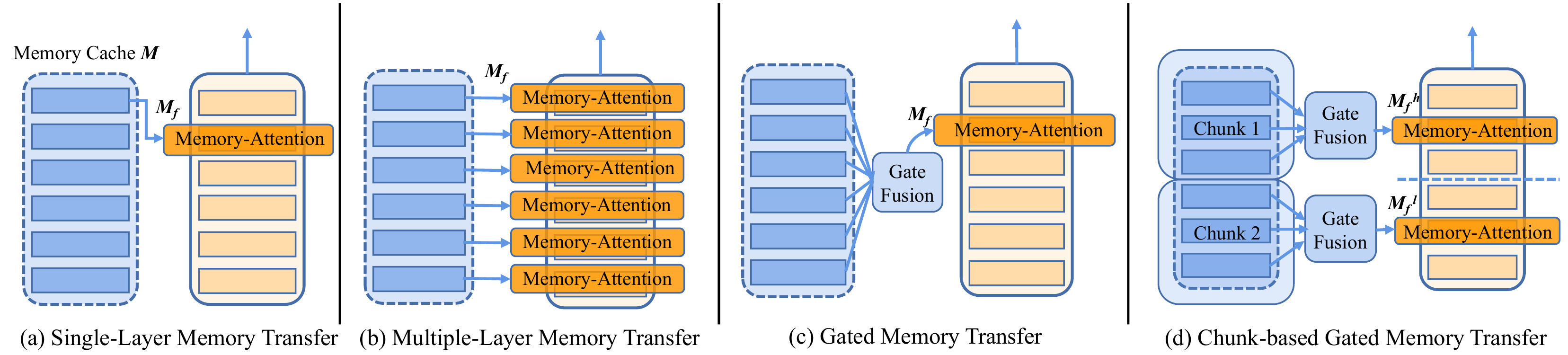}
  \vspace{-0.2in}
  \caption{Memory-augmented strategies of the G-MAP framework. We take a 6-layer model as an example.}
  \vspace{-0.2in}
  \label{fig:architecture}
\end{figure*}

\subsection{Memory-Augmented Layer}
Memory-augmented layer differs from the traditional Transformer layer only in the multi-head self-attention module. The new memory-augmented attention module is proposed to fuse the memory \wan{representation} into the domain-specific PLM, denoted as memory-attention. Its main idea is to linearly transform the memory \wan{representation} into new pairs of (\wan{keys}, values) and concatenate them into the back of pairs produced by the 
domain-specific PLM. Then the multi-head self-attention is performed to adaptively fuse this new concatenated representation. The whole process reuses the parameters of the Transformer layer of domain-specific PLM and does not introduce any new parameters. 

Specifically, if $i$-th Transformer layer is a memory-augmented one, it obtains the domain-specific representation $\boldsymbol{H}_{i-1}$ from the previous layer and \wan{the memory representation} $\boldsymbol{M}_{f}$ as the input and fuses them by the following way: 

\vspace{-3mm}
{\small
% \begin{align}
%  \operatorname{Memory}&\operatorname{-Attention}(\boldsymbol{Q_i}, \boldsymbol{K_i}, \boldsymbol{V_i}, \boldsymbol{M_{f}}) = \notag\\ &\operatorname{Concat}\left(head_{1}, \ldots, head_{k}\right)\boldsymbol{W^o} 
% \end{align}

\begin{align}
 \operatorname{Memory}&\operatorname{-Attention}(\boldsymbol{H}_{i-1}, \boldsymbol{M_{f}}) = \notag\\ &\operatorname{Concat}\left(head_{1}, \ldots, head_{k}\right)\boldsymbol{W^o} 
\end{align}
}
\nobreak
\vspace{-3mm}

\noindent where \wan{$\boldsymbol{M_{f}}$ is the  memory representation directly extracted from the memory cache $\boldsymbol{M}$ or effectively constructed by some adaptive aggregation strategies, which has the same shape as the intermediate hidden state $\boldsymbol{H_{i}}$ of the domain-specific PLM}, $k$ means the number of heads and  $\boldsymbol{W^o}$ is a trainable parameter matrix. Then, $\boldsymbol{M_{f}}$ is linearly transformed into new pairs of (keys, values),  which were appended to the last of domain-specific ones: 

\vspace{-3mm}
{\small
\begin{align}
\boldsymbol{{\tilde{K}_{i,j}}} &= \operatorname{Concat}(\boldsymbol{K_{i,j}}, \boldsymbol{M^{k}_{f}}) \\
\boldsymbol{{\tilde{V}_{i,j}}} &=\operatorname{Concat}(\boldsymbol{V_{i,j}}, \boldsymbol{M^{v}_{f}})\\
 \boldsymbol{Q_{i,j}}, \boldsymbol{K_{i,j}}, &\boldsymbol{V_{i,j}} = \notag \\ &\boldsymbol{H_{i-1}}\boldsymbol{W_{i,j}^{q}}, \boldsymbol{H_{i-1}}\boldsymbol{W_{i,j}^{k}},\boldsymbol{H_{i-1}}\boldsymbol{W_{i,j}^{v}} \\
\boldsymbol{M^{k}_{f}}, \boldsymbol{M^{v}_{f}} &= \boldsymbol{M_{f}}\boldsymbol{W_{i,j}^{k}}, \boldsymbol{M_{f}}\boldsymbol{W_{i,j}^{v}}
\end{align}
}
\nobreak
\vspace{-3mm}

\noindent where $\boldsymbol{W_{i,j}^{q}}$, $\boldsymbol{W_{i,j}^{k}}$ and $\boldsymbol{W_{i,j}^{v}}$ are trainable parameters to generate queries, keys, values respectively, and $j$ refers to $j$-th attention head.  Then the self-attention is performed on the queries and merged pairs of (keys, values) as follows:

\vspace{-2.4mm}
{\small
\begin{align}
  head_{j} &= \operatorname{Softmax}\left(\frac{\boldsymbol{Q_{i,j}} \boldsymbol{{\tilde{K}_{i,j}}^{T}} }{\sqrt{d_{k}}} \right)\boldsymbol{{\tilde{V}_{i,j}}}
\end{align}
}
\nobreak
\vspace{-2.4mm}

\noindent where $d_k$ is the head dimension acting as a scaling factor.  Firstly, a unified attention matrix is computed by the standard scaled dot-product of each query against the keys of general memory and the domain-specific keys. Then, a softmax operation gets the normalized scores that weigh and sum these concatenated values. Without additional parameter updates for the general PLM, domain-specific PLM can dynamically capture useful general knowledge and ignore noisy information through the memory-augmented layer.

%Specifically, the same queries interact with both the two contexts.

% $\operatorname{Softmax}\left(\frac{Q^{\prime} K^{\prime T}}{\sqrt{d_{k}}}+
% \operatorname{Prev}^{\prime}\right) V^{\prime}$we concatenate $\boldsymbol{K_{i}}$ and $\boldsymbol{H_{f}^{K}}$ in the $seqLen$ dimension to generate $K_{concat}$

% $\operatorname{Softmax}\left(\frac{Q^{\prime} K^{\prime T}}{\sqrt{d_{k}}}+
% \operatorname{Prev}^{\prime}\right) V^{\prime}$we concatenate $\boldsymbol{K_{i}}$ and $\boldsymbol{H_{f}^{K}}$ in the $seqLen$ dimension to generate $K_{concat}$

\subsection{Memory-Augmented Strategies}
The remaining problem is how to build the memory \wan{representation} $\boldsymbol{M_{f}}$ from the \wan{memory cache $\boldsymbol{M}$} and which layer of the domain-specific PLM should be the memory-augmented layer to fuse \wan{$\boldsymbol{M_{f}}$}. Essentially, it is a many-to-many layer assignment problem between the general PLM and the domain-specific PLM. To study the effect of layer assignment, we propose and compare different strategies, as shown in Figure~\ref{fig:architecture}.

\hspace*{\fill}

\noindent \textbf{Single-Layer Memory Transfer}
We first consider a single-layer memory transfer approach, where the last hidden state of \wan{the memory cache $\boldsymbol{M}$} is extracted as $\boldsymbol{M_f}$ and then it is fused into one layer of domain-specific PLM with memory-attention. We choose the layer near the top of the domain-specific PLM model as the memory-augmented layer which performs best in the experiment. This strategy does not require additional parameters.

\hspace*{\fill}

\noindent \textbf{Multiple-Layer Memory Transfer}
The single-layer memory transfer may ignore the knowledge learned from shallow layers of the general PLM. To perform layer-wise interaction between the general PLM and the domain-specific PLM, we propose a multiple-layer transfer strategy. This strategy leverages \wan{all hidden states from the memory cache $\boldsymbol{M}$ as the memory representations and then fuses them} into the corresponding layers of the domain-specific PLM, which also does not introduce any new parameters.  

\hspace*{\fill}

\noindent \textbf{Gated Memory Transfer} 
Multiple-layer memory transfer uses the \wan{hidden states} output by all layers of the frozen general PLM as the memory \wan{representations}, which inevitably introduces homogeneous and noisy information. To avoid the problem, we further propose the strategy of gated memory transfer, which firstly exploits the token-level gate mechanism to adaptively weigh and sum representations of different layers into a memory representation, and then it will be fused into one layer of domain-specific PLM. We also choose the layer near the top of the domain-specific PLM as the memory-augmented one which achieves optimal performance in the experiment. The gate fusion mechanism is formulated as below:

% \begin{small}
% \begin{equation}
% \begin{array}{c}

% \boldsymbol{m_{f}^{t}}=\sum_{l=1}^{L}\alpha_{l}^{t}\boldsymbol{m_{l}^{t}}, \\
% \\
% \alpha_{l}^{t}=\frac{\exp \left(g\left( \boldsymbol{m_{l}^{t}}\right)\right)}{\sum_{i=1}^{L}\operatorname{exp} \left(g\left(\boldsymbol{m_{i}^{t}}\right)\right)}, \\
% \\
% \boldsymbol{M_{f}}=\left\{\boldsymbol{m_{f}^{1}}, \boldsymbol{m_{f}^{2}, \ldots, \boldsymbol{m_{f}^{t}}, \ldots, \boldsymbol{m_{f}^{n}}}\right\}
% \end{array} \label{eq:gate_fusion}
% \end{equation}
% \end{small}

\vspace{-3mm}
{\small
\begin{align}
\boldsymbol{m_{f}^{t}} &=\sum_{l=1}^{L}\alpha_{l}^{t}\boldsymbol{m_{l}^{t}} \\
\alpha_{l}^{t} &=\frac{\exp \left(g\left( \boldsymbol{m_{l}^{t}}\right)\right)}{\sum_{i=1}^{L}\operatorname{exp} \left(g\left(\boldsymbol{m_{i}^{t}}\right)\right)}\\
 \boldsymbol{M_{f}} &=\left\{ \boldsymbol{m_{f}^{1}}, \boldsymbol{m_{f}^{2}, \ldots, \boldsymbol{m_{f}^{t}}, \ldots, \boldsymbol{m_{f}^{n}}}\right\} 
\end{align}
}
\nobreak
\vspace{-3mm}

\noindent where $\boldsymbol{m_{l}^{t}}$ is the token representation, $t$ denotes the token index, $l$ is the layer index, \wan{$n$ is the length of tokens} and $g$ refers to a linear layer. We utilize a softmax function to calculate the importance of tokens in different layers. Therefore, the output token representation $\boldsymbol{m_{f}^{t}}$ is obtained by weighing the $t$-th token representations from different layers with their corresponding importance $\alpha_{l}^{t}$. Finally, the built memory \wan{representation} $\boldsymbol{M_{f}}$ is fused to the memory-augmented layer.

\begin{table*}
\centering
\scalebox{0.8}{
\begin{tabular}{l|cccccccc}
\midrule[1.2pt]
 \textbf{Domain} & \multicolumn{2}{c}{ \textbf{BIOMED} } & \multicolumn{2}{c}{ \textbf{CS} } & \multicolumn{2}{c}{ \textbf{NEWS} } & \multicolumn{2}{c}{ \textbf{REVIEWS} }  \\
\midrule
 \textbf{Dataset} & \text{CP} &  \text{RCT} & \text{CI} &  \text{SE} &  \text{HP} &  \text{AG} &  \text{AM} & \text{IMDB} \\

\midrule
\textbf{Fine-Tuning}        & $81.9_{0.1}$ &  $87.2_{0.1}$ & $63.0_{5.8}$ & $77.3_{1.9}$ & $86.6_{0.9}$ &$93.9_{0.2}$ & $65.1_{3.4}$ & $95.0_{0.2}$   \\ 
\textbf{DAPT}              & $84.2_{0.2}$ &  $87.6_{0.1}$ & $75.4_{2.5}$ &  $80.8_{1.5}$ & $88.2_{5.9}$ & $93.9_{0.2}$ & $66.5_{1.4}$ & $95.4_{0.1}$ \\
\textbf{Logits Fusion}     & $84.4_{0.3}$ &  $87.6_{0.1}$ & $77.5_{2.5}$ & $82.9_{1.3}$ & $91.8_{2.5}$ & $93.8_{0.1}$ & $67.6_{0.2}$ & $95.1_{0.2}$ \\
\textbf{Ensemble LMs}      & $84.6_{0.2}$ &  $87.4_{0.2}$ & $76.2_{2.2}$ & $82.6_{1.0}$ & $91.4_{2.8}$ & $93.9_{0.2}$ & $67.5_{1.1}$ & $95.1_{0.2}$ \\
\midrule
\textbf{G-MAP} (\textit{Single-Layer Memory Transfer})           & $84.7_{0.3}$ & $87.8_{0.1}$ & $78.6_{2.3}$ &  $83.2_{0.6}$ & $93.3_{1.6}$ & $94.0_{0.1}$ & $68.5_{0.6}$ & $95.4_{0.1}$ \\
\textbf{G-MAP} (\textit{Multiple-Layer Memory Transfer})         & $84.7_{0.4}$ &  $87.7_{0.1}$ & $74.9_{1.3}$ &  $82.0_{1.1}$ & $92.2_{2.3}$ & $93.8_{0.2}$ & $67.6_{0.6}$ & $95.2_{0.2}$ \\
\textbf{G-MAP} (\textit{Gated Memory Transfer})               & $84.8_{0.2}$ &  $87.8_{0.1}$ & $77.9_{1.5}$ & $83.2_{1.3}$ & $93.2_{2.0}$ & $94.0_{0.2}$ & $68.1_{0.5}$ & $95.4_{0.2}$ \\
\textbf{G-MAP} (\textit{Chunk-based Gated Memory Transfer})   & $\textbf{85.0}_{0.3}$ &  $\textbf{87.9}_{0.2}$ & $\textbf{79.4}_{1.0}$ &  $\textbf{83.8}_{0.8}$ & $\textbf{95.2}_{0.0}$ & $\textbf{94.1}_{0.2}$ & $\textbf{69.0}_{0.8}$ & $\textbf{95.6}_{0.2}$ \\

\midrule[1.2pt]
\end{tabular}
}
\vspace{-1mm}
\caption{
The comparison against baselines on text classification tasks and performance of the proposed memory-augmented strategies. We report the averages across five random seeds, with standard deviations as subscripts. The best performance for each benchmark is marked in black bold. CP, CI, SE, HP, AG and AM denote \textsc{ChemProt}, \textsc{ACL-ARC}, \textsc{SciERC}, \textsc{HyperPartisan}, \textsc{AGNews} and \textsc{Amazon}, respectively.
} 
\vspace{-1mm}
\label{tab:main_table_1}
\end{table*}

\begin{table*}
\centering
\scalebox{0.8}{
\begin{tabular}{l|cccccccc}
\midrule[1.2pt]
 \textbf{Domain} & \multicolumn{2}{c}{ \textbf{BIOMED} } & \multicolumn{2}{c}{ \textbf{CS} } & \multicolumn{2}{c}{ \textbf{NEWS} } & \multicolumn{2}{c}{ \textbf{REVIEWS} }\\
\midrule
 \textbf{Dataset} & \text{CP} &  \text{RCT} & \text{CI} &  \text{SE} &  \text{HP} &  \text{AG} &  \text{AM} & \text{IMDB}  \\

\midrule
\textbf{TAPT}   & $82.6_{0.4}$ &  $87.7_{0.1}$ & $67.4_{1.8}$ &  $79.3_{1.5}$ & $90.4_{5.2}$ & $94.5_{0.1}$ & $68.5_{1.9}$ & $95.5_{0.1}$ \\
\textbf{DAPT+TAPT}   & $84.4_{0.4}$ &  $87.8_{0.1}$ & $75.6_{3.8}$ & $81.3_{1.7}$ & $90.0_{6.6}$ & $94.6_{0.2}$ & $68.7_{1.8}$ & $95.6_{0.1}$ \\
\midrule 
\textbf{G-MAP$^{*}$}  & $\textbf{85.1}_{0.3}$ &  $\textbf{87.9}_{0.0}$ & $\textbf{79.6}_{2.2}$ &  $\textbf{83.9}_{1.1}$ & $\textbf{95.2}_{0.0}$ & $\textbf{94.6}_{0.1}$ & $\textbf{69.9}_{0.3}$ & $\textbf{95.8}_{0.1}$ \\

\midrule[1.2pt]
\end{tabular}
}
\vspace{-1mm}
\caption{
The experimental results of G-MAP compared with TAPT and DAPT+TAPT on domain classification tasks. $^{*}$ means G-MAP framework built with the backbone PLM pre-trained with the process of DAPT then TAPT.
} 
\vspace{-1mm}
\label{tab:main_table_2}
\end{table*}

\hspace*{\fill}

\noindent \textbf{Chunk-based Gated Memory Transfer }
The previous work~\cite{DBLP:conf/naacl/Liu0BPS19,DBLP:conf/blackboxnlp/PhangLB21} have observed that the representations from upper layers and lower layers of pre-trained language models are significantly different. Motivated by this observation, based on the gated memory transfer, we further propose a chunk-based variant, which separates the layers of general PLM into a high-level chunk and a low-level chunk, and then apply the gate fusion strategy to get upper and lower-layer memory \wan{representations} $\boldsymbol{M_{f}^{h}}$ and $\boldsymbol{M_{f}^{l}}$, respectively. Finally, we fuse them into two memory-augmented layers in the domain-specific PLM, and the details of different layer selections for this strategy are presented in Section~\ref{sec:layer_selection_new}.

\section{Experiments}

In this section, we first introduce the evaluation tasks and metrics. Then we illustrate the baseline methods, and implementation settings. Finally, we conduct the experimental analysis of G-MAP.
\subsection{Datasets and Metrics}
\noindent \textbf{Datasets} We evaluate our model on three tasks: text classification, QA, and NER. For text classification, we conduct experiments on eight tasks that cover four domains, including \textsc{ChemProt}~\cite{DBLP:journals/biodb/KringelumKBLOT16} and \textsc{RCT} \cite{Dernoncourt2017PubMed2R} in the biomedical domain, \textsc{ACL-ARC} \cite{Jurgens2018MeasuringTE} and \textsc{SciERC} \cite{Luan2018MultiTaskIO} in the computer science domain, \textsc{HyperPartisan} \cite{Kiesel2019SemEval2019T4} and \textsc{AGNews} \cite{Zhang2015CharacterlevelCN} in the news domain, \textsc{Helpfulness} \cite{McAuley2015ImageBasedRO} and \textsc{IMDB} \cite{Maas2011LearningWV} in the reviews domain. In addition, we 
use micro-F1 as the metric for ChemProt and RCT, and use macro-F1 for the other datasets following \cite{DBLP:conf/acl/GururanganMSLBD20}. 
For NER, we use two datasets, including NCBI-Disease \cite{Dogan2014NCBIDC} in the biomedical domain, CoNNL-2003 \cite{Sang2003IntroductionTT} in the news domain. We use the F1 score as the evaluation metric.
For QA, we utilize two datasets including Medication \cite{Pampari2018emrQAAL} in the biomedical domain, NewsQA \cite{Trischler2017NewsQAAM} in the news domain. We use the Exact-Match (EM) and the F1 score as the evaluation metrics. The detailed description and statistics of each task are shown in Appendix~\ref{sec:dataset_stats}.

\subsection{Baselines}
% One of the baseline method is ensemble learning, which means directly add the output logits of general domain model and domain-specific model.
In our experiment, all the baselines are built on the RoBERTa-base. The details baselines of text classification tasks are described as follows:
\begin{itemize}
   \setlength{\itemsep}{0pt}
   \setlength{\parsep}{0pt}
   \setlength{\parskip}{0pt}
    \item \textbf{Fine-Tuning}: directly fine-tuning the general PLM for downstream tasks.
    \item \textbf{DAPT}~\cite{DBLP:conf/acl/GururanganMSLBD20}: pre-training the general PLM with large-scale domain unlabeled corpora to get the domain-specific PLM then fine-tuning it.
\item \textbf{Logits Fusion}: a straightforward method combines the frozen general PLM and the domain-specific PLM by adding their logits. This method is optimized end-to-end and does not include any memory-augmented strategies in the model.
\item \textbf{Ensemble LMs}: an ensemble method that adds the predicted probabilities of the fine-tuned general and the domain-specific PLMs for final prediction.
    \item \textbf{TAPT}~\cite{DBLP:conf/acl/GururanganMSLBD20}: task-adaptive pretraining continues to pre-train the PLM on the training dataset, and then we fine-tune it for the downstream tasks.
\end{itemize}
For the NER and QA tasks, in addition to these above baselines, we also introduce KALA \cite{DBLP:journals/corr/abs-2204-10555}. \wan{Since KALA is only verified in QA and NER, we use it as a baseline for these two tasks.}
\begin{itemize}
   \setlength{\itemsep}{0pt}
   \setlength{\parsep}{0pt}
   \setlength{\parskip}{0pt}
   \item \textbf{KALA}: constructing an entity memory and knowledge graph on a task-specific domain and then augmenting PLM by this additional knowledge.
\end{itemize}

\subsection{Implementation}
We implement the G-MAP framework based on RoBERTa-base. For the domain-specific PLM, we use the released pre-trained weights DAPT\footnote{https://github.com/allenai/dont-stop-pretraining}. More details on fine-tuning of the downstream tasks are shown in Appendix~\ref{sec:implementation_details}.

%% Question answer and ner
\begin{table*}
\centering
\scalebox{0.8}{
\begin{tabular}{l|cccc}
\midrule[1.2pt]
 \textbf{Domain}  & \multicolumn{2}{c}{\textbf{BIOMED}}& \multicolumn{2}{c}{\textbf{NEWS}} \\
\midrule
 \textbf{Dataset} & \text{Medication (QA)} &  \text{NCBI-Disease (NER)}  & \text{NewsQA (QA)}  & \text{CoNNL-2003 (NER)} \\

\midrule
\textbf{Fine-Tuning}   & $26.9_{0.2}$ | $71.5_{0.5}$ & $86.9_{1.0}$ & $57.2_{0.6}$ | $71.9_{0.4}$     &     $95.6_{0.2}$  \\
\textbf{TAPT}          & $27.0_{0.2}$ | $71.2_{0.3}$ & $86.2_{0.8}$ &  $57.2_{0.5}$ | $71.8_{0.3}$  &      $95.6_{0.3}$                \\ 
\textbf{DAPT}          & $27.2_{0.3}$ | $71.4_{0.4}$ &      $87.2_{0.4}$     &     $58.7_{0.6}$ | $72.4_{0.4}$  &           $95.8_{0.2}$            \\
\textbf{KALA}          & $27.3_{0.4}$ | $71.1_{0.5}$ & $87.7_{0.3}$    & $58.0_{0.6}$ | $72.7_{0.3}$ &    $95.4_{0.3}$   \\
\midrule
\textbf{G-MAP}  & $\textbf{29.1}_{0.3}$ | $\textbf{72.2}_{0.4}$ &  $\textbf{88.7}_{0.2}$ & $\textbf{59.9}_{0.8}$ | $\textbf{73.3}_{0.4}$ & $\textbf{96.2}_{0.2}$\\
\midrule[1.2pt]
\end{tabular}
}
\vspace{-1mm}
\caption{
The experimental results of extractive QA and NER tasks in biomedical and news domains. We use Exact Match and F1 score as the metrics for the QA tasks: Medication and NewsQA, and F1 score for NER tasks: NCBI-Disease and CoNNL-2003. 
} \label{tab:main_table_3}
\vspace{-1mm}
\end{table*}

\subsection{Results and Analysis}
Our experiment results on the domain-specific classification tasks are shown in Table~\ref{tab:main_table_1} and~\ref{tab:main_table_2}, the results of QA and NER tasks are shown on Table~\ref{tab:main_table_3}.

\hspace*{\fill}

\noindent \textbf{Performance on Domain Classification Tasks } 
From Table~\ref{tab:main_table_1}, we can observe that G-MAP with the proposed chuck-based gated memory transfer can achieve better results than all the baselines, which proves that incorporating memory from the general frozen PLM is beneficial for the domain-specific PLM.  Specifically, the strategy of chunk-based gate memory transfer outperforms other strategies, we conjecture that it adaptively selects the token-level information across different layers and adequately utilizes the general knowledge from both the high-level and low-level chunks. However, we also observe that multiple-layer memory transfer has little improvement compared with the baselines, because it incorporates excessive redundant and noisy information from the general PLM without the proposed gated fusion. Besides, single-layer memory transfer is a simple yet effective strategy achieving better than the baselines and the non-gated fusion strategy of multiple-layer memory transfer.

Since the chunk-based gate memory transfer strategy achieves the best performance compared with the baselines, we use it as the default memory-augmented strategy within G-MAP in the following experiments.

\hspace*{\fill}

\vspace{-2mm}

\noindent \textbf{Comparison with Further TAPT }
Further task-adaptive pre-training (TAPT) has been proven to improve the domain-adaptive pre-training (DAPT) \cite{DBLP:conf/acl/GururanganMSLBD20}. To demonstrate the effectiveness of G-MAP on TAPT, we build a G-MAP framework that replaces the domain-specific pre-trained PLM with the task-adaptive pre-trained PLM. From Table~\ref{tab:main_table_2}, we find that our G-MAP also outperforms DAPT+TAPT on all datasets, indicating that the proposed framework is general for different backbone models, including the domain-adaptive and the task-adaptive PLMs.

\hspace*{\fill}

\vspace{-2mm}

\noindent \textbf{Effectiveness for QA and NER }
We also evaluate G-MAP on the tasks of QA and NER, and the experiment results are shown in Table~\ref{tab:main_table_3}. From the results, we see that our method achieves better results than the baselines on all datasets, especially KALA \cite{DBLP:journals/corr/abs-2204-10555}, which spends a considerable effort to construct entity memory and knowledge graph from the contexts. These results further demonstrate the effectiveness of G-MAP. 

\section{Further Discussion}

In the following sections, we conduct some detailed analysis of G-MAP to demonstrate the effectiveness of the frozen general PLM and memory-attention. Moreover, we apply the proposed framework in the pre-training stage and study the effect of layer selection on performance.

\subsection{Effectiveness of Frozen Memory}
% \begin{figure}[htbp]
%   \centering
%   \includegraphics[width=0.9\linewidth]{EMNLP 2022/frozen.pdf}
%   \vspace{-0.1in}
%   \caption{Comparison of utilizing frozen and unfrozen generic models}
%   \vspace{-0.1in}
%   \label{fig:frozen_compared}
% \end{figure}

We compare the frozen and unfrozen ways when building the general memory representation, and the results are shown in Table~\ref{tab:frozen_compared}. From the table, we observe that the frozen method is better than the unfrozen model on all datasets, and both are better than the baseline DAPT. We argue that using the frozen memory has two advantages: (1) more efficient in model training without updating the parameters of the general PLM; (2) keeps general knowledge from PLM unchanged when fine-tuning, so it does not lead to a forgetting problem.  

\begin{table}
\centering
\scalebox{0.75}{
\begin{tabular}{l|cccc}
\midrule[1.2pt]
 \textbf{Domain}  & \multicolumn{2}{c}{\textbf{BIOMED}}& \multicolumn{2}{c}{\textbf{CS}} \\
\midrule
 \textbf{Dataset} & \text{CP} &  \text{RCT}  & \text{CI}  & \text{SE} \\
\midrule
\textbf{DAPT} & $84.2_{0.2}$ &  $87.6_{0.1}$ & $75.4_{2.5}$  & $80.8_{1.5}$ \\
\midrule
\textbf{G-MAP w/o Frozen }   & $84.3_{0.5}$ & $87.3_{0.3}$ & $78.1_{2.2}$ & $82.4_{1.1}$ \\ 
\textbf{G-MAP}          & $\textbf{85.0}_{0.3}$  & $\textbf{87.9}_{0.2}$ &  $\textbf{79.4}_{1.0}$  &  $\textbf{83.8}_{0.8}$               \\

\midrule[1.2pt]
\end{tabular}
}
\caption{Results of utilizing the frozen and unfrozen general PLMs. } \label{tab:frozen_compared}
\end{table}

\subsection{Effectiveness of Memory-Attention} 

\begin{table}
\centering
\scalebox{0.75}{
\begin{tabular}{l|cccc}
\midrule[1.2pt]
 \textbf{Domain}  & \multicolumn{2}{c}{\textbf{BIOMED}}& \multicolumn{2}{c}{\textbf{CS}} \\
\midrule
 \textbf{Dataset} & \text{CP} &  \text{RCT}  & \text{CI}  & \text{SE} \\
\midrule
\textbf{Cross-Attention}    & $82.8_{0.4}$ &  $86.6_{0.1}$ & $73.5_{2.0}$  & $82.2_{1.0}$ \\
\textbf{Gate-Attention}     & $84.7_{0.3}$ & $87.6_{0.1}$ & $78.1_{1.4}$ & $83.4_{0.7}$ \\ 
\midrule
\textbf{Memory-Attention}   & $\textbf{85.0}_{0.3}$  & $\textbf{87.9}_{0.2}$ &  $\textbf{79.4}_{1.0}$  &  $\textbf{83.8}_{0.8}$               \\

\midrule[1.2pt]
\end{tabular}
}
\caption{Performance of using the proposed memory-attention and other main-stream attention-based variants in G-MAP framework. We evaluate them on the datasets of biomedical and cs domains.} \label{tab:attention_variants}
\end{table}

To study the effectiveness of our proposed memory-attention module, we compare it with other attention-based variants within a memory-augmented layer. Specifically, cross-attention is an attention module widely-used in the multi-modal learning \cite{DBLP:conf/nips/LiSGJXH21, DBLP:journals/corr/abs-2111-08276}, we apply it to adaptively fuse the memory representation $\boldsymbol{M_{f}}$ and the output representation from the self-attention module. We also include the gate-attention \cite{DBLP:journals/corr/abs-2203-08913} as the fusion baseline, which utilizes a gate mechanism to weigh and sum the local and external memory for long-sequence modeling. As shown in Table~\ref{tab:attention_variants}, our memory-attention module outperforms other variants without additional trainable parameters.

\subsection{Layer Selection for Memory-Attention}
\label{sec:layer_selection_new}

\begin{figure}[htbp]
  \centering
  \includegraphics[width=0.95\linewidth]{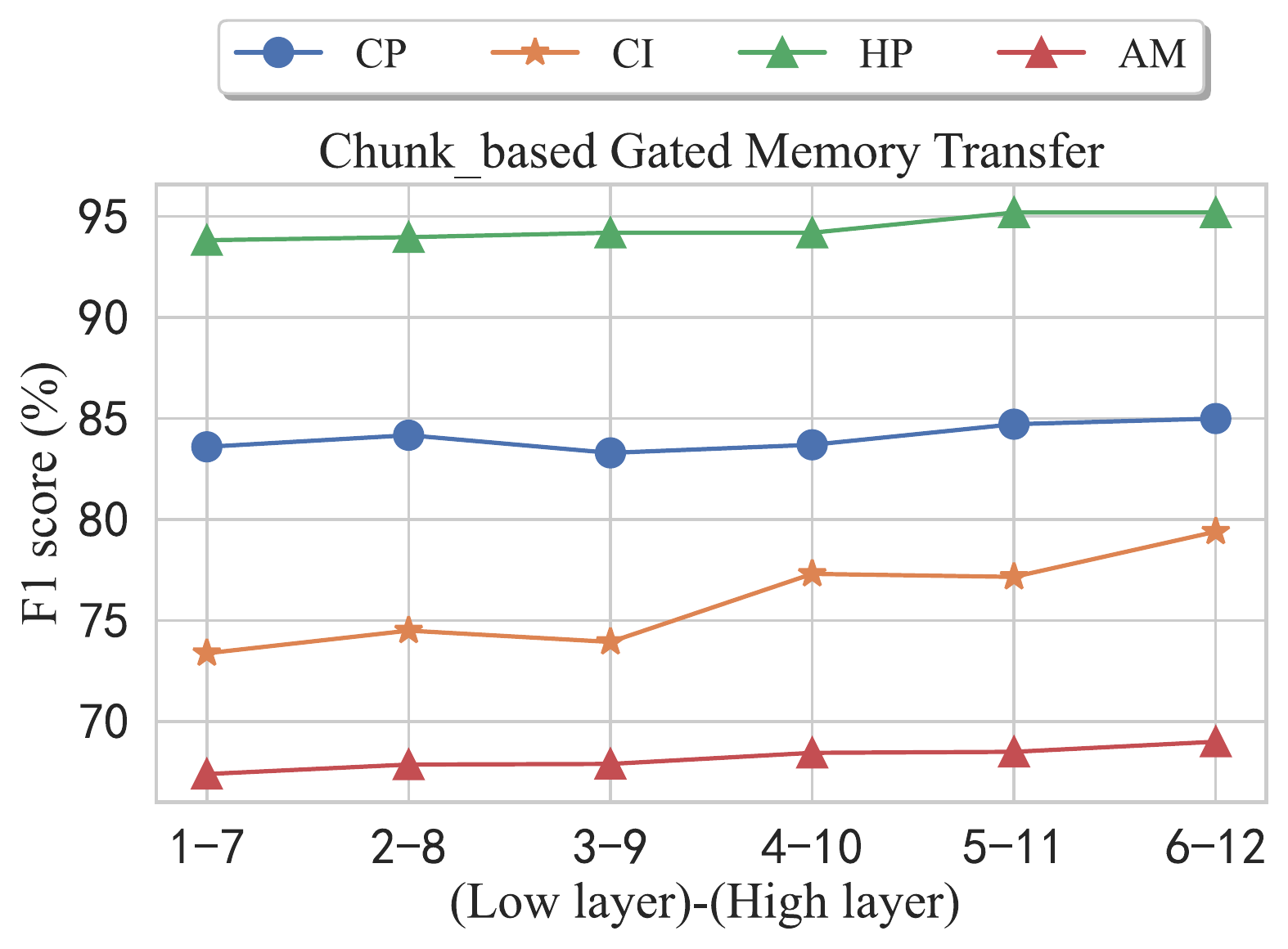}
  \vspace{-0.1in}
  \caption{Performance of different layer selections in chunk-based gate memory transfer strategy.}
  \vspace{-0.1in}
  \label{fig:layer_selection}

\end{figure}

Besides the strategy of multiple-layer memory transfer, other strategies need to do the layer selection. For the strategies of single and gated-memory transfer, we fuse the memory representation $\boldsymbol{M_{f}}$ into different layers $\left\{ 3, 6, 9, 12 \right\}$ in the domain-specific PLM and find that the layer 9 as the memory-augmented layer can achieve the best performance in both strategies. We present more detailed results in Appendix~\ref{sec:layer_selection}. For the chunk-based gate memory transfer strategy, we experiment with transferring the memory representation of the high-level chunk into layers 7 to 12, and the other one of the low-level chunk into layers 1 to 6, which keeps the same layer interval between the two memory-augmented layers in the domain-specific PLM. The experimental results are shown in Figure~\ref{fig:layer_selection}. The results show that there is an increasing tendency when placing memory-augmented layers to the top of the domain-specific PLM. Finally, we choose layers 6 and 12 as the memory-augmented ones for the strategy of chunk-based gated memory transfer.

\begin{figure}[tbp]
\centering
\begin{minipage}[c]{0.16\textwidth}
\centering
\includegraphics[width=2.6cm,height = 2.6cm]{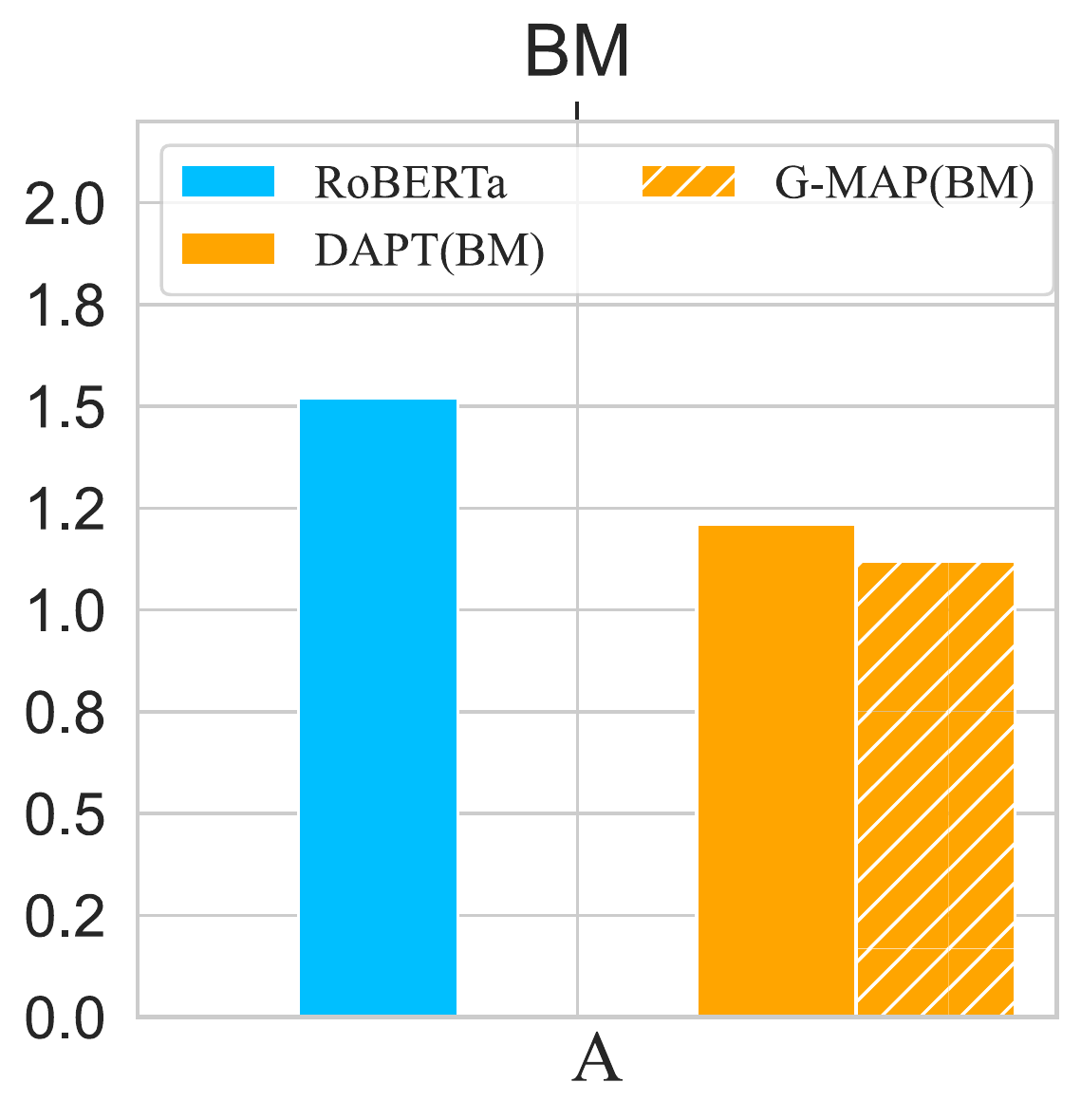}
\end{minipage}
\begin{minipage}[c]{0.15\textwidth}
\centering
\includegraphics[width=2.5cm,height = 2.6cm]{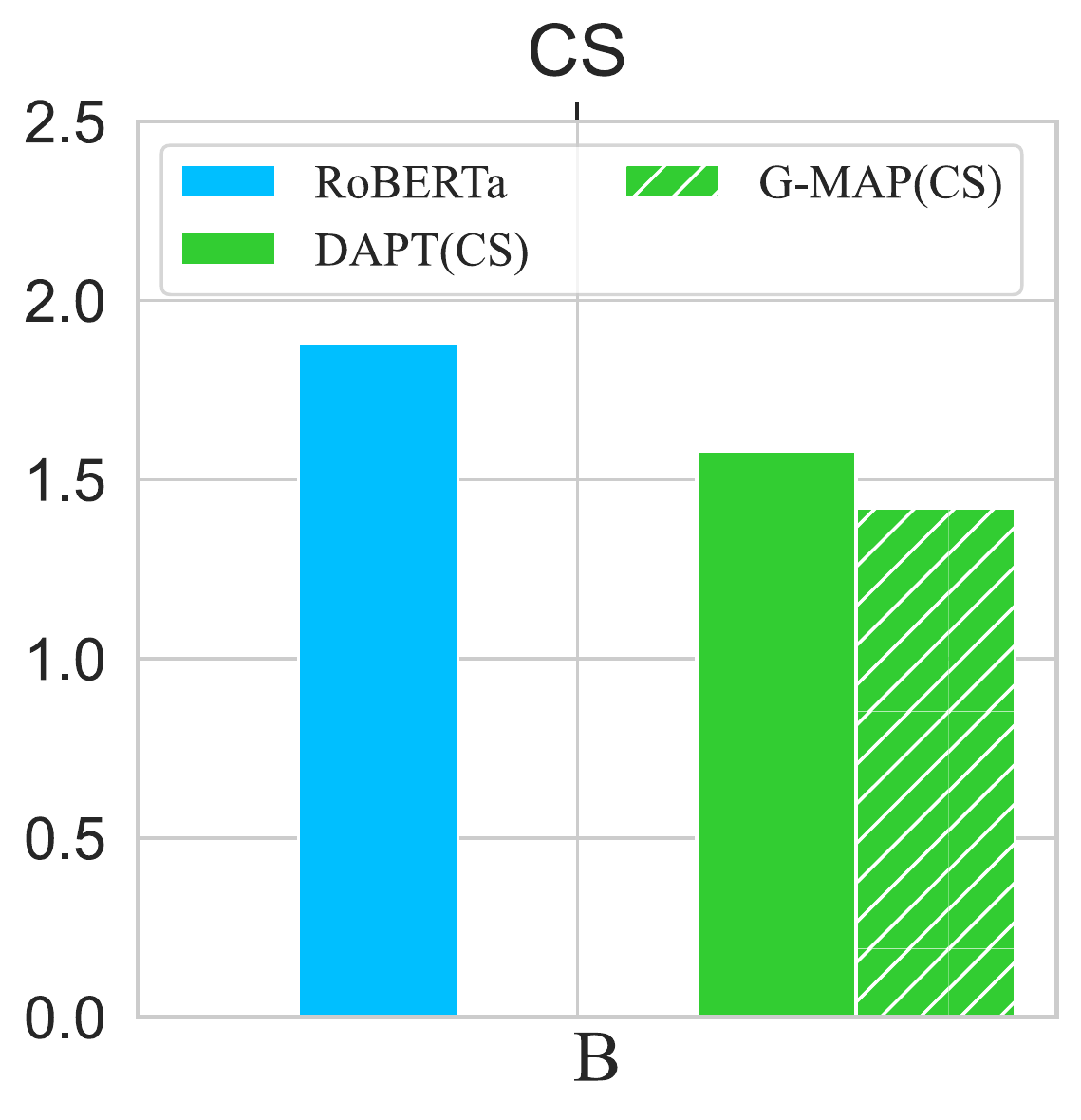}
\end{minipage}
\begin{minipage}[c]{0.15\textwidth}
\centering
\includegraphics[width=2.6cm,height = 2.6cm]{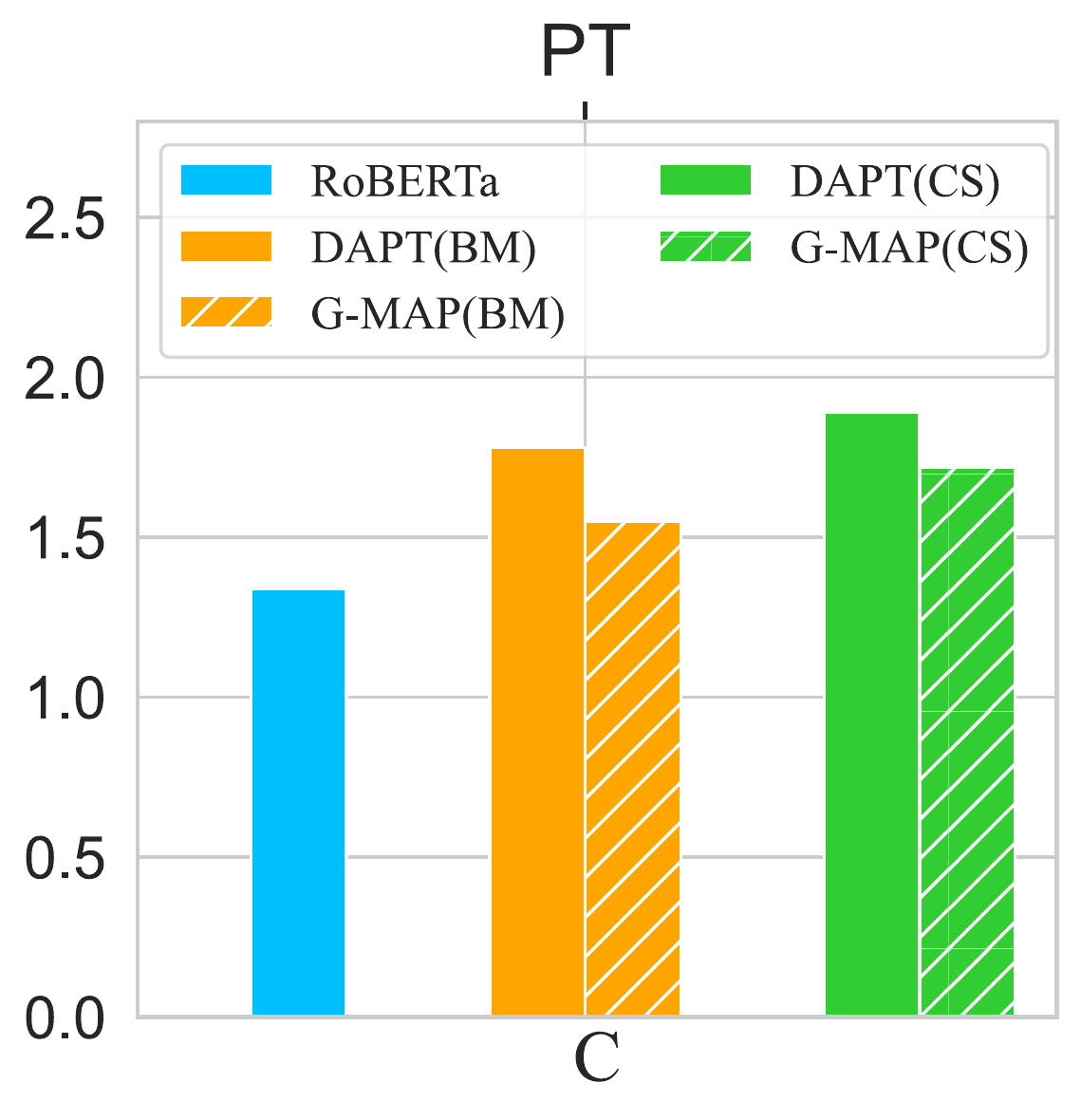}
\end{minipage}
\caption{Maksed LM loss for the pre-training stage (a lower value is better). PT denotes samples similar to RoBERTa's pre-training corpus. DAPT(BM) denotes the domain-specific PLM for the biomedical domain. G-MAP(BM) denotes the G-MAP framework with the biomedical-domain backbone.  For instance, figure A represents further pre-training of the models on the biomedical pre-training samples and then inferring their MLM loss on the test samples.}\label{fig:MLM_loss_pretrained}
\vspace{-0.2in}
\end{figure}

\subsection{Apply G-MAP in the Pre-training Stage}
\label{sec:adaptive_pretrain}
In the previous experiment, we incorporated the domain-specific PLM with the G-MAP framework in the fine-tuning stage. In this section, we further study whether G-MAP is beneficial for the pre-training stage. To this end, we randomly sample 50k documents from general\footnote{https://github.com/soskek/bookcorpus}, biomedical and computer science domains \cite{Lo2020S2ORCTS}, respectively. In addition, we randomly split 70$\%$ of the data from each domain as the pre-training samples and the rest data as the test samples. More details about pre-processing the domain samples are shown in Appendix~\ref{sec:implementation_details}.  Then, we pre-train the models on the pre-training samples and calculate the masked LM loss on the test samples. From the experiment results shown in Figure~\ref{fig:MLM_loss_pretrained}, compared with baseline DAPT, we observe that utilizing G-MAP can reduce masked LM loss on the biomedical, cs, and general domains. These results demonstrate that the proposed G-MAP also mitigates catastrophic forgetting during the adaptive pre-training.

% \begin{figure}[htbp]
%   \centering
%   \includegraphics[width=0.85\linewidth]{heatmap_2.pdf}
%   \vspace{-0.1in}
%   \caption{Maksed LM loss for cross-domain (lower value implies better fit). PT denotes the samples similar to RoBERTa's pre-training corpus. For instance, DAPT(BM) denotes the domain-adaptive pre-trained model for biomedical domain. DAPT(BM)+MAP denotes combining DAPT(BM) with MAP for further in-domain or cross-domain adaptive pre-training.}
  
%   \vspace{-0.1in}
%   \label{fig:cross_domain_mlm}

% \end{figure}

\section{Related work}

\noindent \textbf{Domain Adaptation for PLMs }  
Recently, the domain shift problem of PLMs has attracted increasing research \cite{ DBLP:conf/emnlp/BeltagyLC19, DBLP:journals/corr/abs-1904-05342, DBLP:journals/bioinformatics/LeeYKKKSK20, DBLP:conf/acl/GururanganMSLBD20, wan2024med} since the domain discrepancies between the pre-training corpora and the downstream tasks can lead to a significant performance drop. To bridge the domain gap, SciBERT \cite{DBLP:conf/emnlp/BeltagyLC19} and BioBERT \cite{DBLP:journals/bioinformatics/LeeYKKKSK20} further pre-train BERT with 1.14M scientific papers from Semantic Scholar corpus and biomedical documents, respectively, which can improve the performance of domain-specific NLU tasks compared with general BERT. Also, \citet{DBLP:conf/acl/GururanganMSLBD20} proposed domain-adaptive pre-training (DAPT) and task-adaptive pre-training (TAPT). Concretely, DAPT continues to pre-train the PLM with domain-specific corpora, while TAPT directly pre-trains the PLM on the task dataset. Moreover, BT-TAPT \cite{DBLP:journals/corr/abs-2107-10474} inherits the crucial step of TAPT and leverages the back-translated strategy to augment the task data to improve the performance of PLM. TAPTER \cite{DBLP:conf/acl/NishidaNY21} equips TAPT with domain-specific word embedding regularization to improve fine-tuning performance. However, the above approaches suffer from catastrophic forgetting of general domain knowledge after adaptive pre-training, which leads to sub-optimal performance on downstream tasks.

\hspace*{\fill} 

\vspace{-1mm}

\noindent \textbf{Catastrophic Forgetting } 
\emph{Catastrophic forgetting} is a common phenomenon for continual learning, and it occurs when a training model forgets previously learned knowledge and over-fits to new tasks \cite{mccloskey:catastrophic}. Typically, regularization-based methods \cite{DBLP:journals/corr/GoodfellowMDCB13, DBLP:journals/corr/KirkpatrickPRVD16, DBLP:conf/icml/SerraSMK18, deng2021flattening} exploit regularization to constrain the parameter update to alleviate the forgetting problem, and \wan{the memory-based methods \cite{DBLP:conf/nips/GuoLYR20, DBLP:conf/iclr/SahaG021} mitigate forgetting by storing important samples from past tasks in the external memory and rehearsing them via some gradient 
transformation strategies.} In addition, plenty of works have been proposed to address \emph{catastrophic forgetting} for NLP tasks.  \citet{Dakwale} subtly minimized KL-divergence of prediction losses as a regularization term between fine-tuning and general domain models. \citet{Cheolhyoung} introduced a new regularization technique to mix the PLM parameters with vanilla parameters instead of stochastical dropout. \citet{DBLP:conf/emnlp/ChenHCCLY20} adopted multi-task learning to jointly learn pre-training and downstream tasks with less forgetting during fine-tuning. \citet{DBLP:conf/acl/Xie0G020} preserved the model neurons of general and language-specific parts during fine-tuning. However, our method is orthogonal to the above approaches since we aim to effectively incorporate the domain-specific PLM with the memory representation built from the frozen general PLM to solve the forgetting issue without adding additional regularization terms in the model or using external memory to preserve samples from the past tasks.

\hspace*{\fill} 

\vspace{-1mm}

\noindent \textbf{Knowledge-Enhanced PLMs } Knowledge-enhanced methods have shown effectiveness for PLMs via introducing internal or external knowledge. To improve the performance of fine-tuning tasks, REINA \cite{DBLP:conf/acl/WangXFLSX0022} retrieves the labeled training instances most similar to the input data and concatenates them before feeding them into PLMs. Besides, RETRO \cite{DBLP:journals/corr/abs-2112-04426} enhances the auto-regressive language model via leveraging a pre-trained frozen BERT model to retrieve related texts and then use a chunked cross-attention module to incorporate them. Memorizing transformer \cite{DBLP:journals/corr/abs-2203-08913} \wan{leverages a learned gate to combine the attention results of the local context and the external context retrieved from previously seen sub-sequences.} KALA \cite{DBLP:journals/corr/abs-2204-10555} is the approach most relevant to our work. It incorporates intermediate hidden representations with domain-specific entities and their relational facts during task-specific fine-tuning for \wan{domain tasks}. However, our method doesn't need to retrieve similar texts or construct additional knowledge graphs. We propose several memory-augmented strategies to build the memory representation and then transfer it into the domain-specific PLM to mitigate the forgetting of general knowledge.

\vspace{-1mm}

\section{Conclusion}
In this work, we propose G-MAP, a novel framework that utilizes the memory-augmented layer to fuse the memory representation built from the frozen general PLM to mitigate \emph{catastrophic forgetting} of general knowledge caused by domain-adaptive pre-training. We explore different memory-augmented strategies to construct the memory representation and empirically find that chunk-based gate memory transfer achieves the most optimal performance. We validate G-MAP on various domains of classification, QA, and NER tasks. The results show that our method consistently outperforms existing baselines on all datasets, implying that explicitly leveraging forgotten general knowledge is beneficial for domain-specific downstream tasks.

\section{Limitations}

% One limitation of our MAP framework is that it cannot intelligently choose which layer of the domain-specific PLM to use as the memory-augmented layer in the training process depending on the task, since our proposed memory-augmented strategies only experimentally explore choosing different layers in the domain-specific PLM as the memory-augmented layers to obtain the optimal strategy. We leave the problem that how to adaptively transfer the memory representation to the specific layer of domain-specific PLM during training as the future work. 

% Second, we have not validated the effectiveness of our MAP framework during large-scale domain pre-training. Currently, our MAP framework has only been validated in domain downstream tasks and very small-scale domain pre-training experiment in sec~\ref{sec:adaptive_pretrain}. We also leave this challenge as our future work.

% Finally, we only validate the performance of MAP framework on RoBERTa architecture, which is a contextual-based pre-trained language model. We have not explore the feasibility of utilizing our method on generative models like GPT \cite{radford2018improving} or BART \cite{DBLP:conf/acl/LewisLGGMLSZ20}. Considering that generalizing our framework to generative models is also valuable for research, we leave it for future exploration. 

Our G-MAP framework has been validated on domain-specific tasks and a small-scale domain pre-training experiment in Section~\ref{sec:adaptive_pretrain}. Due to the lack of large GPU resource, we have not validated our G-MAP framework in large-scale pre-training, a more challenging setting that we leave as future work. We also consider automatic layer selection to be an under-studied problem and believe that AutoML techniques~\cite{DBLP:conf/icml/PhamGZLD18, DBLP:conf/icml/TanL19}, such as evolutionary search~\cite{DBLP:journals/tec/DebAPM02, DBLP:journals/tsmc/ChenCPJ21, liang2020large, liang2020many}, will be promising methods. Finally, the proposed framework is built on the encoder-only model, RoBERTa-base. In the future, we will apply our framework on the other types of architectures~\cite{wang2024iot, wan2023text}, such as decoder-only GPT~\cite{radford2018improving, wan2023efficient} and encoder-decoder BART~\cite{DBLP:conf/acl/LewisLGGMLSZ20}, or other domains, such as math~\cite{cobbe2021training, xiong2022self}, recommendation~\cite{wan2023spatio}, and medical signal~\cite{liu2023etp}.
\section*{Acknowledge}
This work is supported by National Key R$\& $D Program of China(2022YFE0200700), National Natural Science Foundation of China (Project No. 62006219) and Natural Science Foundation of Guangdong Province (2022A1515011579).
\bibliography{anthology,custom}

\begin{thebibliography}{62}
\expandafter\ifx\csname natexlab\endcsname\relax\def\natexlab#1{#1}\fi

\bibitem[{Beltagy et~al.(2019)Beltagy, Lo, and
  Cohan}]{DBLP:conf/emnlp/BeltagyLC19}
Iz~Beltagy, Kyle Lo, and Arman Cohan. 2019.
\newblock \href {https://arxiv.org/abs/1903.10676} {Scibert: {A} pretrained
  language model for scientific text}.
\newblock In \emph{{EMNLP}}.

\bibitem[{Borgeaud et~al.(2021)Borgeaud, Mensch, Hoffmann, Cai, Rutherford,
  Millican, van~den Driessche, Lespiau, Damoc, Clark, de~Las~Casas, Guy,
  Menick, Ring, Hennigan, Huang, Maggiore, Jones, Cassirer, Brock, Paganini,
  Irving, Vinyals, Osindero, Simonyan, Rae, Elsen, and
  Sifre}]{DBLP:journals/corr/abs-2112-04426}
Sebastian Borgeaud, Arthur Mensch, Jordan Hoffmann, Trevor Cai, Eliza
  Rutherford, Katie Millican, George van~den Driessche, Jean{-}Baptiste
  Lespiau, Bogdan Damoc, Aidan Clark, Diego de~Las~Casas, Aurelia Guy, Jacob
  Menick, Roman Ring, Tom Hennigan, Saffron Huang, Loren Maggiore, Chris Jones,
  Albin Cassirer, Andy Brock, Michela Paganini, Geoffrey Irving, Oriol Vinyals,
  Simon Osindero, Karen Simonyan, Jack~W. Rae, Erich Elsen, and Laurent Sifre.
  2021.
\newblock \href {http://arxiv.org/abs/2112.04426} {Improving language models by
  retrieving from trillions of tokens}.
\newblock \emph{CoRR}, abs/2112.04426.

\bibitem[{Chen et~al.(2021)Chen, Cheng, Pedrycz, and
  Jin}]{DBLP:journals/tsmc/ChenCPJ21}
Huangke Chen, Ran Cheng, Witold Pedrycz, and Yaochu Jin. 2021.
\newblock \href {https://doi.org/10.1109/TSMC.2019.2930737} {Solving
  many-objective optimization problems via multistage evolutionary search}.
\newblock \emph{{IEEE} Trans. Syst. Man Cybern. Syst.}

\bibitem[{Chen et~al.(2020)Chen, Hou, Cui, Che, Liu, and
  Yu}]{DBLP:conf/emnlp/ChenHCCLY20}
Sanyuan Chen, Yutai Hou, Yiming Cui, Wanxiang Che, Ting Liu, and Xiangzhan Yu.
  2020.
\newblock \href {https://arxiv.org/pdf/2004.12651.pdf} {Recall and learn:
  Fine-tuning deep pretrained language models with less forgetting}.
\newblock In \emph{{EMNLP}}.

\bibitem[{Cobbe et~al.(2021)Cobbe, Kosaraju, Bavarian, Chen, Jun, Kaiser,
  Plappert, Tworek, Hilton, Nakano et~al.}]{cobbe2021training}
Karl Cobbe, Vineet Kosaraju, Mohammad Bavarian, Mark Chen, Heewoo Jun, Lukasz
  Kaiser, Matthias Plappert, Jerry Tworek, Jacob Hilton, Reiichiro Nakano,
  et~al. 2021.
\newblock Training verifiers to solve math word problems.
\newblock \emph{arXiv preprint arXiv:2110.14168}.

\bibitem[{Dakwale and Monz.(2017)}]{Dakwale}
Praveen Dakwale and Christof Monz. 2017.
\newblock \href
  {https://staff.science.uva.nl/c.monz/ltl/publications/mtsummit2017.pdf}
  {Fine-tuning for neural machine translation with limited degradation across
  in-and out-of-domain data}.
\newblock In \emph{Proceedings of the XVI Machine Translation Summit}.

\bibitem[{Deb et~al.(2002)Deb, Agrawal, Pratap, and
  Meyarivan}]{DBLP:journals/tec/DebAPM02}
Kalyanmoy Deb, Samir Agrawal, Amrit Pratap, and T.~Meyarivan. 2002.
\newblock \href {https://doi.org/10.1109/4235.996017} {A fast and elitist
  multiobjective genetic algorithm: {NSGA-II}}.
\newblock \emph{{IEEE} Trans. Evol. Comput.}

\bibitem[{Deng et~al.(2021)Deng, Chen, Hao, Wang, and
  Heng}]{deng2021flattening}
Danruo Deng, Guangyong Chen, Jianye Hao, Qiong Wang, and Pheng-Ann Heng. 2021.
\newblock \href
  {https://proceedings.neurips.cc/paper/2021/hash/9b16759a62899465ab21e2e79d2ef75c-Abstract.html}
  {Flattening sharpness for dynamic gradient projection memory benefits
  continual learning}.
\newblock \emph{Advances in Neural Information Processing Systems},
  34:18710--18721.

\bibitem[{Dernoncourt and Lee(2017)}]{Dernoncourt2017PubMed2R}
Franck Dernoncourt and Ji~Young Lee. 2017.
\newblock \href {https://aclanthology.org/I17-2052/} {Pubmed 200k rct: a
  dataset for sequential sentence classification in medical abstracts}.
\newblock In \emph{IJCNLP}.

\bibitem[{Devlin et~al.(2019)Devlin, Chang, Lee, and
  Toutanova}]{DBLP:conf/naacl/DevlinCLT19}
Jacob Devlin, Ming{-}Wei Chang, Kenton Lee, and Kristina Toutanova. 2019.
\newblock \href {https://arxiv.org/abs/1810.04805} {{BERT:} pre-training of
  deep bidirectional transformers for language understanding}.
\newblock In \emph{{NAACL-HLT}}.

\bibitem[{Dogan et~al.(2014)Dogan, Leaman, and Lu}]{Dogan2014NCBIDC}
Rezarta~Islamaj Dogan, Robert Leaman, and Zhiyong Lu. 2014.
\newblock \href {https://pubmed.ncbi.nlm.nih.gov/24393765/} {Ncbi disease
  corpus: A resource for disease name recognition and concept normalization}.
\newblock \emph{Journal of biomedical informatics}.

\bibitem[{Goodfellow et~al.(2014)Goodfellow, Mirza, Da, Courville, and
  Bengio}]{DBLP:journals/corr/GoodfellowMDCB13}
Ian~J. Goodfellow, Mehdi Mirza, Xia Da, Aaron~C. Courville, and Yoshua Bengio.
  2014.
\newblock \href {https://arxiv.org/pdf/1312.6211.pdf} {An empirical
  investigation of catastrophic forgeting in gradient-based neural networks}.
\newblock In \emph{{ICLR}}.

\bibitem[{Guo et~al.(2020)Guo, Liu, Yang, and Rosing}]{DBLP:conf/nips/GuoLYR20}
Yunhui Guo, Mingrui Liu, Tianbao Yang, and Tajana Rosing. 2020.
\newblock \href
  {https://proceedings.neurips.cc/paper/2020/hash/0b5e29aa1acf8bdc5d8935d7036fa4f5-Abstract.html}
  {Improved schemes for episodic memory-based lifelong learning}.
\newblock In \emph{{NIPS}}.

\bibitem[{Gururangan et~al.(2020)Gururangan, Marasovic, Swayamdipta, Lo,
  Beltagy, Downey, and Smith}]{DBLP:conf/acl/GururanganMSLBD20}
Suchin Gururangan, Ana Marasovic, Swabha Swayamdipta, Kyle Lo, Iz~Beltagy, Doug
  Downey, and Noah~A. Smith. 2020.
\newblock Don't stop pretraining: Adapt language models to domains and tasks.
\newblock In \emph{{ACL}}.

\bibitem[{Huang et~al.(2019)Huang, Altosaar, and
  Ranganath}]{DBLP:journals/corr/abs-1904-05342}
Kexin Huang, Jaan Altosaar, and Rajesh Ranganath. 2019.
\newblock \href {http://arxiv.org/abs/1904.05342} {Clinicalbert: Modeling
  clinical notes and predicting hospital readmission}.
\newblock \emph{CoRR}, abs/1904.05342.

\bibitem[{Jurgens et~al.(2018)Jurgens, Kumar, Hoover, McFarland, and
  Jurafsky}]{Jurgens2018MeasuringTE}
David Jurgens, Srijan Kumar, Raine Hoover, Daniel~A. McFarland, and Dan
  Jurafsky. 2018.
\newblock \href {https://aclanthology.org/Q18-1028.pdf} {Measuring the
  evolution of a scientific field through citation frames}.
\newblock \emph{Transactions of the Association for Computational Linguistics}.

\bibitem[{Kang et~al.(2022)Kang, Baek, and
  Hwang}]{DBLP:journals/corr/abs-2204-10555}
Minki Kang, Jinheon Baek, and Sung~Ju Hwang. 2022.
\newblock \href {https://arxiv.org/abs/2204.10555} {{KALA:} knowledge-augmented
  language model adaptation}.
\newblock In \emph{{NAACL}}.

\bibitem[{Kiesel et~al.(2019)Kiesel, Mestre, Shukla, Vincent, Adineh, Corney,
  Stein, and Potthast}]{Kiesel2019SemEval2019T4}
Johannes Kiesel, Maria Mestre, Rishabh Shukla, Emmanuel Vincent, Payam Adineh,
  D.~Corney, Benno Stein, and Martin Potthast. 2019.
\newblock \href {https://aclanthology.org/S19-2145/} {Semeval-2019 task 4:
  Hyperpartisan news detection}.
\newblock In \emph{*SEMEVAL}.

\bibitem[{Kirkpatrick et~al.(2016)Kirkpatrick, Pascanu, Rabinowitz, Veness,
  Desjardins, Rusu, Milan, Quan, Ramalho, Grabska{-}Barwinska, Hassabis,
  Clopath, Kumaran, and Hadsell}]{DBLP:journals/corr/KirkpatrickPRVD16}
James Kirkpatrick, Razvan Pascanu, Neil~C. Rabinowitz, Joel Veness, Guillaume
  Desjardins, Andrei~A. Rusu, Kieran Milan, John Quan, Tiago Ramalho, Agnieszka
  Grabska{-}Barwinska, Demis Hassabis, Claudia Clopath, Dharshan Kumaran, and
  Raia Hadsell. 2016.
\newblock \href {http://arxiv.org/abs/1612.00796} {Overcoming catastrophic
  forgetting in neural networks}.
\newblock \emph{CoRR}, abs/1612.00796.

\bibitem[{Kringelum et~al.(2016)Kringelum, Kj{\ae}rulff, Brunak, Lund, Oprea,
  and Taboureau}]{DBLP:journals/biodb/KringelumKBLOT16}
Jens Kringelum, Sonny~Kim Kj{\ae}rulff, S{\o}ren Brunak, Ole Lund, Tudor~I.
  Oprea, and Olivier Taboureau. 2016.
\newblock \href {https://pubmed.ncbi.nlm.nih.gov/26876982/} {Chemprot-3.0: a
  global chemical biology diseases mapping}.
\newblock \emph{Database J. Biol. Databases Curation}.

\bibitem[{Lee et~al.(2020{\natexlab{a}})Lee, Cho, and Kang}]{Cheolhyoung}
Cheolhyoung Lee, Kyunghyun Cho, and Wanmo Kang. 2020{\natexlab{a}}.
\newblock \href {https://openreview.net/pdf?id=HkgaETNtDB} {Mixout: Effective
  regularization to fine-tune large-scale pretrained language models.}
\newblock In \emph{{ICLR}}.

\bibitem[{Lee et~al.(2020{\natexlab{b}})Lee, Yoon, Kim, Kim, Kim, So, and
  Kang}]{DBLP:journals/bioinformatics/LeeYKKKSK20}
Jinhyuk Lee, Wonjin Yoon, Sungdong Kim, Donghyeon Kim, Sunkyu Kim, Chan~Ho So,
  and Jaewoo Kang. 2020{\natexlab{b}}.
\newblock \href {https://arxiv.org/ftp/arxiv/papers/1901/1901.08746.pdf}
  {Biobert: a pre-trained biomedical language representation model for
  biomedical text mining}.
\newblock \emph{Bioinform.}

\bibitem[{Lee et~al.(2021)Lee, Kim, and
  Kang}]{DBLP:journals/corr/abs-2107-10474}
Junghoon Lee, Jounghee Kim, and Pilsung Kang. 2021.
\newblock \href {https://arxiv.org/abs/2107.10474} {Back-translated task
  adaptive pretraining: Improving accuracy and robustness on text
  classification}.
\newblock \emph{CoRR}, abs/2107.10474.

\bibitem[{Levine et~al.(2022)Levine, Dalmedigos, Ram, Zeldes, Jannai, Muhlgay,
  Osin, Lieber, Lenz, Shalev{-}Shwartz, Shashua, Leyton{-}Brown, and
  Shoham}]{Levine}
Yoav Levine, Itay Dalmedigos, Ori Ram, Yoel Zeldes, Daniel Jannai, Dor Muhlgay,
  Yoni Osin, Opher Lieber, Barak Lenz, Shai Shalev{-}Shwartz, Amnon Shashua,
  Kevin Leyton{-}Brown, and Yoav Shoham. 2022.
\newblock \href {https://doi.org/10.48550/arXiv.2204.10019} {Standing on the
  shoulders of giant frozen language models}.
\newblock volume abs/2204.10019.

\bibitem[{Lewis et~al.(2020)Lewis, Liu, Goyal, Ghazvininejad, Mohamed, Levy,
  Stoyanov, and Zettlemoyer}]{DBLP:conf/acl/LewisLGGMLSZ20}
Mike Lewis, Yinhan Liu, Naman Goyal, Marjan Ghazvininejad, Abdelrahman Mohamed,
  Omer Levy, Veselin Stoyanov, and Luke Zettlemoyer. 2020.
\newblock \href {https://arxiv.org/abs/1910.13461} {{BART:} denoising
  sequence-to-sequence pre-training for natural language generation,
  translation, and comprehension}.
\newblock In \emph{{ACL}}.

\bibitem[{Li et~al.(2021)Li, Selvaraju, Gotmare, Joty, Xiong, and
  Hoi}]{DBLP:conf/nips/LiSGJXH21}
Junnan Li, Ramprasaath~R. Selvaraju, Akhilesh Gotmare, Shafiq~R. Joty, Caiming
  Xiong, and Steven~Chu{-}Hong Hoi. 2021.
\newblock \href
  {https://proceedings.neurips.cc/paper/2021/hash/505259756244493872b7709a8a01b536-Abstract.html}
  {Align before fuse: Vision and language representation learning with momentum
  distillation}.
\newblock In \emph{{NIPS}}.

\bibitem[{Li and Hoiem(2016)}]{DBLP:conf/eccv/LiH16}
Zhizhong Li and Derek Hoiem. 2016.
\newblock \href {https://arxiv.org/abs/1606.09282} {Learning without
  forgetting}.
\newblock In \emph{{ECCV}}.

\bibitem[{Liang et~al.(2020{\natexlab{a}})Liang, Li, and Wan}]{liang2020large}
Zhenyu Liang, Yunfan Li, and Zhongwei Wan. 2020{\natexlab{a}}.
\newblock Large scale many-objective optimization driven by distributional
  adversarial networks.
\newblock \emph{arXiv preprint arXiv:2003.07013}.

\bibitem[{Liang et~al.(2020{\natexlab{b}})Liang, Li, and Wan}]{liang2020many}
Zhenyu Liang, Yunfan Li, and Zhongwei Wan. 2020{\natexlab{b}}.
\newblock Many-objective estimation of distribution optimization algorithm
  based on wgan-gp.
\newblock \emph{arXiv preprint arXiv:2003.08295}.

\bibitem[{Liu et~al.(2023)Liu, Wan, Cheng, Zhang, and Arcucci}]{liu2023etp}
Che Liu, Zhongwei Wan, Sibo Cheng, Mi~Zhang, and Rossella Arcucci. 2023.
\newblock Etp: Learning transferable ecg representations via ecg-text
  pre-training.
\newblock \emph{arXiv preprint arXiv:2309.07145}.

\bibitem[{Liu et~al.(2019{\natexlab{a}})Liu, Gardner, Belinkov, Peters, and
  Smith}]{DBLP:conf/naacl/Liu0BPS19}
Nelson~F. Liu, Matt Gardner, Yonatan Belinkov, Matthew~E. Peters, and Noah~A.
  Smith. 2019{\natexlab{a}}.
\newblock \href {https://doi.org/10.18653/v1/n19-1112} {Linguistic knowledge
  and transferability of contextual representations}.
\newblock In \emph{{NAACL-HLT}}, pages 1073--1094.

\bibitem[{Liu et~al.(2019{\natexlab{b}})Liu, Ott, Goyal, Du, Joshi, Chen, Levy,
  Lewis, Zettlemoyer, and Stoyanov}]{DBLP:journals/corr/abs-1907-11692}
Yinhan Liu, Myle Ott, Naman Goyal, Jingfei Du, Mandar Joshi, Danqi Chen, Omer
  Levy, Mike Lewis, Luke Zettlemoyer, and Veselin Stoyanov. 2019{\natexlab{b}}.
\newblock \href {http://arxiv.org/abs/1907.11692} {Roberta: {A} robustly
  optimized {BERT} pretraining approach}.
\newblock \emph{CoRR}, abs/1907.11692.

\bibitem[{Lo et~al.(2020)Lo, Wang, Neumann, Kinney, and Weld}]{Lo2020S2ORCTS}
Kyle Lo, Lucy~Lu Wang, Mark Neumann, Rodney~Michael Kinney, and Daniel~S. Weld.
  2020.
\newblock \href {https://arxiv.org/abs/1911.02782} {S2orc: The semantic scholar
  open research corpus}.
\newblock In \emph{{ACL}}.

\bibitem[{Luan et~al.(2018)Luan, He, Ostendorf, and
  Hajishirzi}]{Luan2018MultiTaskIO}
Yi~Luan, Luheng He, Mari Ostendorf, and Hannaneh Hajishirzi. 2018.
\newblock \href {https://aclanthology.org/D18-1360.pdf} {Multi-task
  identification of entities, relations, and coreference for scientific
  knowledge graph construction}.
\newblock In \emph{{EMNLP}}.

\bibitem[{Maas et~al.(2011)Maas, Daly, Pham, Huang, Ng, and
  Potts}]{Maas2011LearningWV}
Andrew~L. Maas, Raymond~E. Daly, Peter~T. Pham, Dan Huang, A.~Ng, and
  Christopher Potts. 2011.
\newblock \href
  {https://ai.stanford.edu/~ang/papers/acl11-WordVectorsSentimentAnalysis.pdf}
  {Learning word vectors for sentiment analysis}.
\newblock In \emph{{ACL}}.

\bibitem[{McAuley et~al.(2015)McAuley, Targett, Shi, and van~den
  Hengel}]{McAuley2015ImageBasedRO}
Julian McAuley, Christopher Targett, Qinfeng Shi, and Anton van~den Hengel.
  2015.
\newblock \href {https://arxiv.org/pdf/1506.04757.pdf} {Image-based
  recommendations on styles and substitutes}.
\newblock \emph{{SIGIR}}.

\bibitem[{Mccloskey and Cohen(1989)}]{mccloskey:catastrophic}
Michael Mccloskey and Neil~J. Cohen. 1989.
\newblock \href
  {https://www.sciencedirect.com/science/article/abs/pii/S0079742108605368}
  {Catastrophic interference in connectionist networks: {T}he sequential
  learning problem}.
\newblock \emph{The Psychology of Learning and Motivation}.

\bibitem[{Neumann et~al.()Neumann, King, Beltagy, and
  Ammar}]{DBLP:conf/bionlp/NeumannKBA19}
Mark Neumann, Daniel King, Iz~Beltagy, and Waleed Ammar.
\newblock \href {https://doi.org/10.18653/v1/w19-5034} {Scispacy: Fast and
  robust models for biomedical natural language processing}.
\newblock In \emph{{BioNLP@ACL}}.

\bibitem[{Nishida et~al.(2021)Nishida, Nishida, and
  Yoshida}]{DBLP:conf/acl/NishidaNY21}
Kosuke Nishida, Kyosuke Nishida, and Sen Yoshida. 2021.
\newblock \href {https://arxiv.org/abs/2109.08354} {Task-adaptive pre-training
  of language models with word embedding regularization}.
\newblock In \emph{Findings of {ACL}}.

\bibitem[{Pampari et~al.(2018)Pampari, Raghavan, Liang, and
  Peng}]{Pampari2018emrQAAL}
Anusri Pampari, Preethi Raghavan, Jennifer~J. Liang, and Jian Peng. 2018.
\newblock \href {https://aclanthology.org/D18-1258/} {emrqa: A large corpus for
  question answering on electronic medical records}.
\newblock In \emph{EMNLP}.

\bibitem[{Pham et~al.(2018)Pham, Guan, Zoph, Le, and
  Dean}]{DBLP:conf/icml/PhamGZLD18}
Hieu Pham, Melody~Y. Guan, Barret Zoph, Quoc~V. Le, and Jeff Dean. 2018.
\newblock \href {http://proceedings.mlr.press/v80/pham18a.html} {Efficient
  neural architecture search via parameter sharing}.
\newblock In \emph{{ICML}}, Proceedings of Machine Learning Research.

\bibitem[{Phang et~al.(2021)Phang, Liu, and
  Bowman}]{DBLP:conf/blackboxnlp/PhangLB21}
Jason Phang, Haokun Liu, and Samuel~R. Bowman. 2021.
\newblock \href {https://aclanthology.org/2021.blackboxnlp-1.42} {Fine-tuned
  transformers show clusters of similar representations across layers}.
\newblock In \emph{BlackboxNLP@EMNLP}.

\bibitem[{Radford et~al.(2018)Radford, Narasimhan, Salimans, and
  Sutskever}]{radford2018improving}
Alec Radford, Karthik Narasimhan, Tim Salimans, and Ilya Sutskever. 2018.
\newblock Improving language understanding by generative pre-training.

\bibitem[{Saha et~al.(2021)Saha, Garg, and Roy}]{DBLP:conf/iclr/SahaG021}
Gobinda Saha, Isha Garg, and Kaushik Roy. 2021.
\newblock \href {https://openreview.net/forum?id=3AOj0RCNC2} {Gradient
  projection memory for continual learning}.
\newblock In \emph{{ICLR}}.

\bibitem[{Sang and Meulder(2003)}]{Sang2003IntroductionTT}
Erik Tjong~Kim Sang and Fien~De Meulder. 2003.
\newblock \href {https://aclanthology.org/W03-0419.pdf} {Introduction to the
  conll-2003 shared task: Language-independent named entity recognition}.
\newblock In \emph{CoNLL}.

\bibitem[{Serr{\`{a}} et~al.()Serr{\`{a}}, Suris, Miron, and
  Karatzoglou}]{DBLP:conf/icml/SerraSMK18}
Joan Serr{\`{a}}, Didac Suris, Marius Miron, and Alexandros Karatzoglou.
\newblock \href {http://proceedings.mlr.press/v80/serra18a.html} {Overcoming
  catastrophic forgetting with hard attention to the task}.
\newblock In \emph{{ICML}}.

\bibitem[{Tan and Le(2019)}]{DBLP:conf/icml/TanL19}
Mingxing Tan and Quoc~V. Le. 2019.
\newblock \href {http://proceedings.mlr.press/v97/tan19a.html} {Efficientnet:
  Rethinking model scaling for convolutional neural networks}.
\newblock In \emph{{ICML}}, Proceedings of Machine Learning Research.

\bibitem[{Thompson et~al.(2019)Thompson, Gwinnup, Khayrallah, Duh, and
  Koehn}]{DBLP:conf/naacl/ThompsonGKDK19}
Brian Thompson, Jeremy Gwinnup, Huda Khayrallah, Kevin Duh, and Philipp Koehn.
  2019.
\newblock \href {https://aclanthology.org/N19-1209.pdf} {Overcoming
  catastrophic forgetting during domain adaptation of neural machine
  translation}.
\newblock In \emph{{NAACL-HLT}}.

\bibitem[{Trischler et~al.(2017)Trischler, Wang, Yuan, Harris, Sordoni,
  Bachman, and Suleman}]{Trischler2017NewsQAAM}
Adam Trischler, Tong Wang, Xingdi Yuan, Justin Harris, Alessandro Sordoni,
  Philip Bachman, and Kaheer Suleman. 2017.
\newblock \href {https://arxiv.org/abs/1611.09830} {Newsqa: A machine
  comprehension dataset}.
\newblock In \emph{Rep4NLP@ACL}.

\bibitem[{Vaswani et~al.(2017)Vaswani, Shazeer, Parmar, Uszkoreit, Jones,
  Gomez, Kaiser, and Polosukhin}]{DBLP:conf/nips/VaswaniSPUJGKP17}
Ashish Vaswani, Noam Shazeer, Niki Parmar, Jakob Uszkoreit, Llion Jones,
  Aidan~N. Gomez, Lukasz Kaiser, and Illia Polosukhin. 2017.
\newblock \href {https://arxiv.org/abs/1706.03762} {Attention is all you need}.
\newblock In \emph{{NIPS}}.

\bibitem[{Wan(2023)}]{wan2023text}
Zhongwei Wan. 2023.
\newblock Text classification: A perspective of deep learning methods.
\newblock \emph{arXiv preprint arXiv:2309.13761}.

\bibitem[{Wan et~al.(2024)Wan, Liu, Zhang, Fu, Wang, Cheng, Ma,
  Quilodr{\'a}n-Casas, and Arcucci}]{wan2024med}
Zhongwei Wan, Che Liu, Mi~Zhang, Jie Fu, Benyou Wang, Sibo Cheng, Lei Ma,
  C{\'e}sar Quilodr{\'a}n-Casas, and Rossella Arcucci. 2024.
\newblock Med-unic: Unifying cross-lingual medical vision-language pre-training
  by diminishing bias.
\newblock \emph{Advances in Neural Information Processing Systems}, 36.

\bibitem[{Wan et~al.(2023{\natexlab{a}})Wan, Liu, Wang, Qiu, Li, Guo, Chen, and
  Wang}]{wan2023spatio}
Zhongwei Wan, Xin Liu, Benyou Wang, Jiezhong Qiu, Boyu Li, Ting Guo, Guangyong
  Chen, and Yang Wang. 2023{\natexlab{a}}.
\newblock Spatio-temporal contrastive learning-enhanced gnns for session-based
  recommendation.
\newblock \emph{ACM Transactions on Information Systems}, 42(2):1--26.

\bibitem[{Wan et~al.(2023{\natexlab{b}})Wan, Wang et~al.}]{wan2023efficient}
Zhongwei Wan, Xin Wang, et~al. 2023{\natexlab{b}}.
\newblock Efficient large language models: A survey.
\newblock \emph{arXiv preprint arXiv:2312.03863}.

\bibitem[{Wang et~al.(2022)Wang, Xu, Fang, Liu, Sun, Xu, Zhu, and
  Zeng}]{DBLP:conf/acl/WangXFLSX0022}
Shuohang Wang, Yichong Xu, Yuwei Fang, Yang Liu, Siqi Sun, Ruochen Xu,
  Chenguang Zhu, and Michael Zeng. 2022.
\newblock \href {https://arxiv.org/pdf/2203.08773.pdf} {Training data is more
  valuable than you think: {A} simple and effective method by retrieving from
  training data}.
\newblock In \emph{{ACL}}.

\bibitem[{Wang et~al.(2024)Wang, Wan, Hekmati, Zong, Alam, Zhang, and
  Krishnamachari}]{wang2024iot}
Xin Wang, Zhongwei Wan, Arvin Hekmati, Mingyu Zong, Samiul Alam, Mi~Zhang, and
  Bhaskar Krishnamachari. 2024.
\newblock Iot in the era of generative ai: Vision and challenges.
\newblock \emph{arXiv preprint arXiv:2401.01923}.

\bibitem[{Wolf et~al.(2020)Wolf, Debut, Sanh, Chaumond, Delangue, Moi, Cistac,
  Rault, Louf, Funtowicz, Davison, Shleifer, von Platen, Ma, Jernite, Plu, Xu,
  Scao, Gugger, Drame, Lhoest, and Rush}]{DBLP:conf/emnlp/WolfDSCDMCRLFDS20}
Thomas Wolf, Lysandre Debut, Victor Sanh, Julien Chaumond, Clement Delangue,
  Anthony Moi, Pierric Cistac, Tim Rault, R{\'{e}}mi Louf, Morgan Funtowicz,
  Joe Davison, Sam Shleifer, Patrick von Platen, Clara Ma, Yacine Jernite,
  Julien Plu, Canwen Xu, Teven~Le Scao, Sylvain Gugger, Mariama Drame, Quentin
  Lhoest, and Alexander~M. Rush. 2020.
\newblock \href {https://doi.org/10.18653/v1/2020.emnlp-demos.6} {Transformers:
  State-of-the-art natural language processing}.
\newblock In \emph{{EMNLP}}.

\bibitem[{Wu et~al.(2022)Wu, Rabe, Hutchins, and
  Szegedy}]{DBLP:journals/corr/abs-2203-08913}
Yuhuai Wu, Markus~N. Rabe, DeLesley Hutchins, and Christian Szegedy. 2022.
\newblock \href {https://doi.org/10.48550/arXiv.2203.08913} {Memorizing
  transformers}.
\newblock \emph{CoRR}, abs/2203.08913.

\bibitem[{Xie et~al.(2021)Xie, Feng, Gu, and Yu}]{DBLP:conf/acl/Xie0G020}
Wanying Xie, Yang Feng, Shuhao Gu, and Dong Yu. 2021.
\newblock \href {https://arxiv.org/pdf/2107.06569.pdf} {Importance-based neuron
  allocation for multilingual neural machine translation}.
\newblock In \emph{{ACL}}.

\bibitem[{Xiong et~al.(2022)Xiong, Wan, Hu, Yang, and Li}]{xiong2022self}
Jing Xiong, Zhongwei Wan, Xiping Hu, Min Yang, and Chengming Li. 2022.
\newblock Self-consistent reasoning for solving math word problems.
\newblock \emph{arXiv preprint arXiv:2210.15373}.

\bibitem[{Zeng et~al.(2021)Zeng, Zhang, and
  Li}]{DBLP:journals/corr/abs-2111-08276}
Yan Zeng, Xinsong Zhang, and Hang Li. 2021.
\newblock \href {https://arxiv.org/abs/2111.08276} {Multi-grained vision
  language pre-training: Aligning texts with visual concepts}.
\newblock In \emph{{ICML}}.

\bibitem[{Zhang et~al.(2015)Zhang, Zhao, and LeCun}]{Zhang2015CharacterlevelCN}
Xiang Zhang, Junbo~Jake Zhao, and Yann LeCun. 2015.
\newblock \href {https://arxiv.org/pdf/1509.01626.pdf} {Character-level
  convolutional networks for text classification}.
\newblock \emph{ArXiv}, abs/1509.01626.

\end{thebibliography}
\bibliographystyle{acl_natbib}

\newpage
\appendix

\section{Dataset Descriptions and Statistics}
\label{sec:dataset_stats}

This section describes the details and statistics of three tasks: domain classification, domain extractive question answering (QA), and named entity recognition (NER). 

\textbf{For the text classification tasks}, we leverage the following datasets covering four domains, including biomedical science, computer science, news, and reviews.
In the biomedical domain, \textsc{ChemProt} \cite{DBLP:journals/biodb/KringelumKBLOT16} is the relation classification dataset based on chemical-protein interaction. \textsc{RCT} \cite{Dernoncourt2017PubMed2R} is the role classification task constructed from the abstract of the biomedical articles. In the computer science domain, \textsc{ACL-ARC} \cite{Jurgens2018MeasuringTE} is the task of annotated citations for articles' functions. \textsc{SciERC} \cite{Luan2018MultiTaskIO} is constructed from scientific abstracts annotated with relation. In the news domain, \textsc{HyperPartisan} \cite{Kiesel2019SemEval2019T4} is the news text classification for determining partisan leanings. \textsc{AGNews} \cite{Zhang2015CharacterlevelCN} is the topic classification for news. In the reviews domain, \textsc{Amazon} \cite{McAuley2015ImageBasedRO} is a binary classification task consisting of feedback on products. \textsc{IMDB} \cite{Maas2011LearningWV} consists of movies reviews, which is a binary sentiment classification dataset.

\textbf{For the NER tasks}, we use two datasets involving the news and biomedical domain. Concretely, CoNLL-2003 \cite{Sang2003IntroductionTT} consists of news stories from the Reuters Corpus. NCBI-Disease \cite{Dogan2014NCBIDC} is annotated with disease mentions.

\textbf{For the QA tasks}, we utilize two domain-specific datasets. Specifically, NewsQA \cite{Trischler2017NewsQAAM} is a machine comprehension dataset consisting of news articles. Medication \cite{Pampari2018emrQAAL} is constructed by electronic medical records from clinical text. 

The detailed statistics of text classification tasks and NER tasks are shown in Table~\ref{tab:statistic_data}, QA in Table~\ref{tab:statistic_data_2}.

\section{Implementation Details}
\label{sec:implementation_details}
We use the huggingface \cite{DBLP:conf/emnlp/WolfDSCDMCRLFDS20} library to implement our G-MAP framework, which contains various transformer-based pre-trained language models (PLMs) and their saved checkpoints. We implement the DAPT, TAPT, and DAPT+TAPT models of biomedical, cs, news, and reviews domains \footnote{https://huggingface.co/allenai} from the library released by \cite{DBLP:conf/acl/GururanganMSLBD20}. All the experiments are implemented on Nvidia V100 GPUs with 32GB memory. We select the best checkpoint on the validation set during training to infer the test set.

\hspace*{\fill}

\vspace{-2mm}

\noindent \textbf{Configurations for Classification Tasks }
In this section, we explain the setting of fine-tuning for domain-specific classification tasks. We fine-tune the domain-specific PLM with our G-MAP framework for 5 to 15 epochs, respectively, with the same learning rate of 4e-5 and the dropout rate of 0.5. The default classification layer of the model is 1 except for the IMDB dataset with 2, and the default maximum sequence length is 256 except for IMDB with 512. We leverage the Adam optimizer to schedule the learning rate, with the Adam epsilon of 1e-8, the Adam beta-1 of 0.9, and the Adam beta-2 of 0.999. We apply the grid-search method to find the optimal batch size and numbers of GPUs for all the classification datasets. The detailed settings are shown in Table~\ref{tab:statistic_data}.

\hspace*{\fill}

\vspace{-2mm}

\noindent \textbf{Configurations for QA and NER}
For the extractive QA tasks, we fine-tune the domain-specific PLM with our G-MAP framework for 3 epochs, which can converge to optimal performance. Besides, we train the model with a maximum sequence length of 384 and the learning rate of 3e-5, the weight decay rate with 1e-2 and the warm-up rate of 6e-2. For the experiments on NER tasks, we fine-tune NCBI-Disease for 20 epochs and CoNNL-2003 for 15 epochs, with the same maximum sequence length of 128, the learning rate of 5e-5, and the weight decay rate and warm-up rate are set to 0. Different from domain classification tasks, we utilize AdamW as the learning rate optimizer instead of Adam. In addition, we also adopt the grid-search method to find the optimal batch size and number of GPUs for all tasks. The detailed settings are shown in Table~\ref{tab:config_qa_ner}.

\hspace*{\fill}

\begin{table} 
\centering
\scalebox{0.75}{
\begin{tabular}{lc}
\midrule[1.2pt]
\textbf{Hyperparameters} & Domain corpus \\

\midrule
Training epochs &  5 to 10 \\
Batch size per GPU     & 32   \\
Number of GPUs  & 4   \\
Maximum Sequence Length & 512   \\
The number of text lines (pre-training) & 35K \\
The number of text lines (inference) & 15K\\
Learning Rate & 4e-5 \\
Learning Rate Optimizer & Adam  \\
Adam epsilon & 1e-8  \\
Adam beta 1 & 0.9  \\
Adam beta 2 & 0.999  \\
\midrule[1.2pt]
\end{tabular}
}
\caption{Hyperparameters for simple adaptive pre-training with G-MAP framework on biomedical and cs corpora. 
} \label{tab:config_adaptive_pretrain}
\end{table}

\vspace{-2mm}
\noindent \textbf{Configurations for small-scale Pre-training  }
This part describes the experimental settings of adaptive pre-training with G-MAP framework. Our simple pre-training experiment needs some external domain-relative corpora of two domains: Biomedical and Computer Science. Following by \cite{DBLP:conf/acl/GururanganMSLBD20}, we adopt \textsc{scispaCy} \cite{DBLP:conf/bionlp/NeumannKBA19} as a sentence splitting tool to obtain abstract and body paragraphs from S2ORC \cite{Lo2020S2ORCTS}. After pre-processing for the corpora, we randomly sample 50K data for each domain and split 70$\%$ of them as pre-training sets and 30$\%$ as test sets. For the general corpus similar to \textsc{RoBERTa}'s pre-training corpus, we also randomly sample 50K data from \textsc{BookCorpus}\footnote{https://github.com/soskek/bookcorpu} and split them by using the method mentioned above. The detailed hyper-parameter settings for this cross-domain adaptive pre-training are shown in Table~\ref{tab:config_adaptive_pretrain}.

\begin{figure}[tbp]
\centering
\begin{minipage}[c]{0.24\textwidth}
\centering
\includegraphics[width=3.8cm,height = 3.8cm]{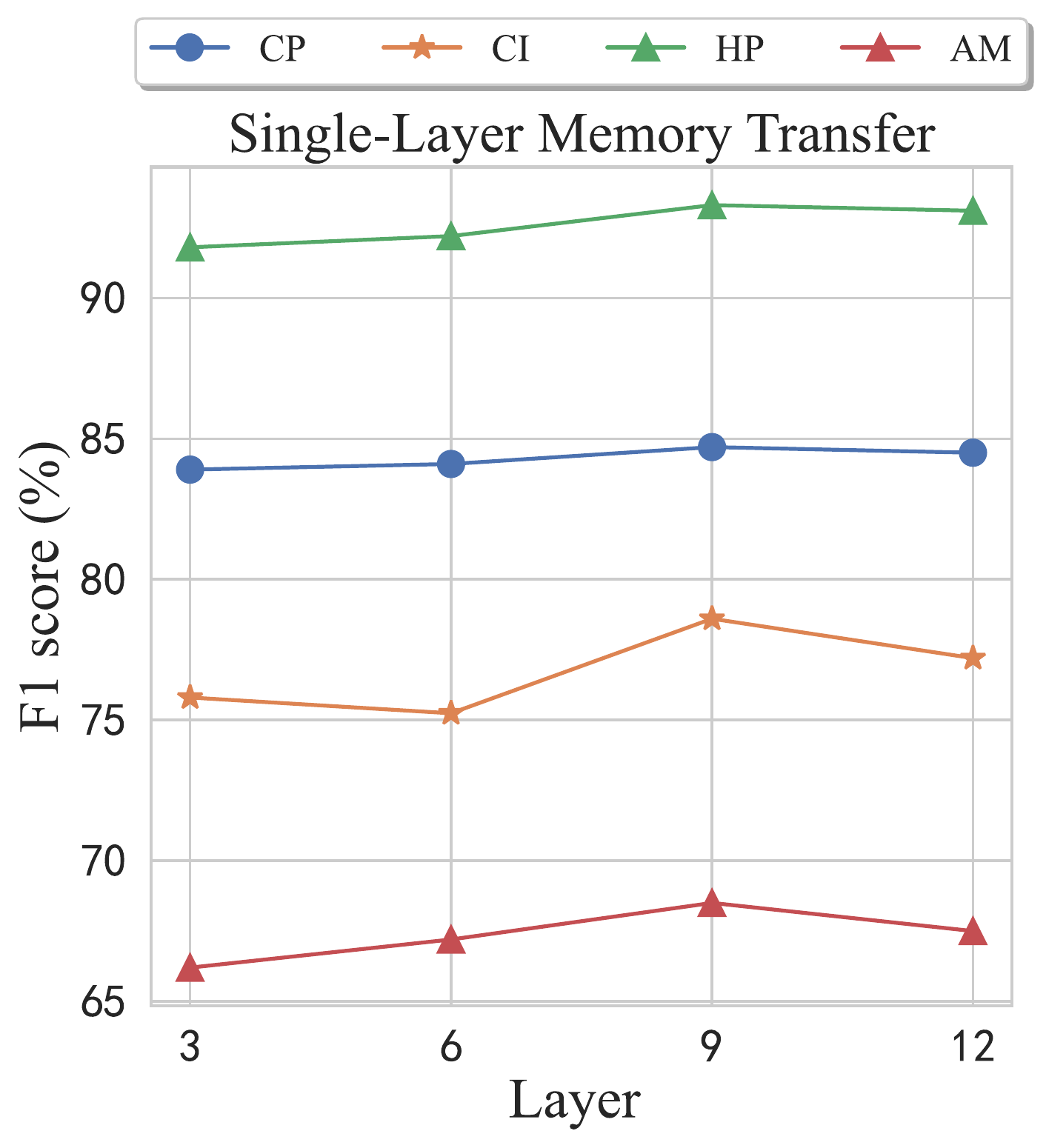}
\end{minipage}%
\begin{minipage}[c]{0.24\textwidth}
\centering
\includegraphics[width=3.8cm,height = 3.8cm]{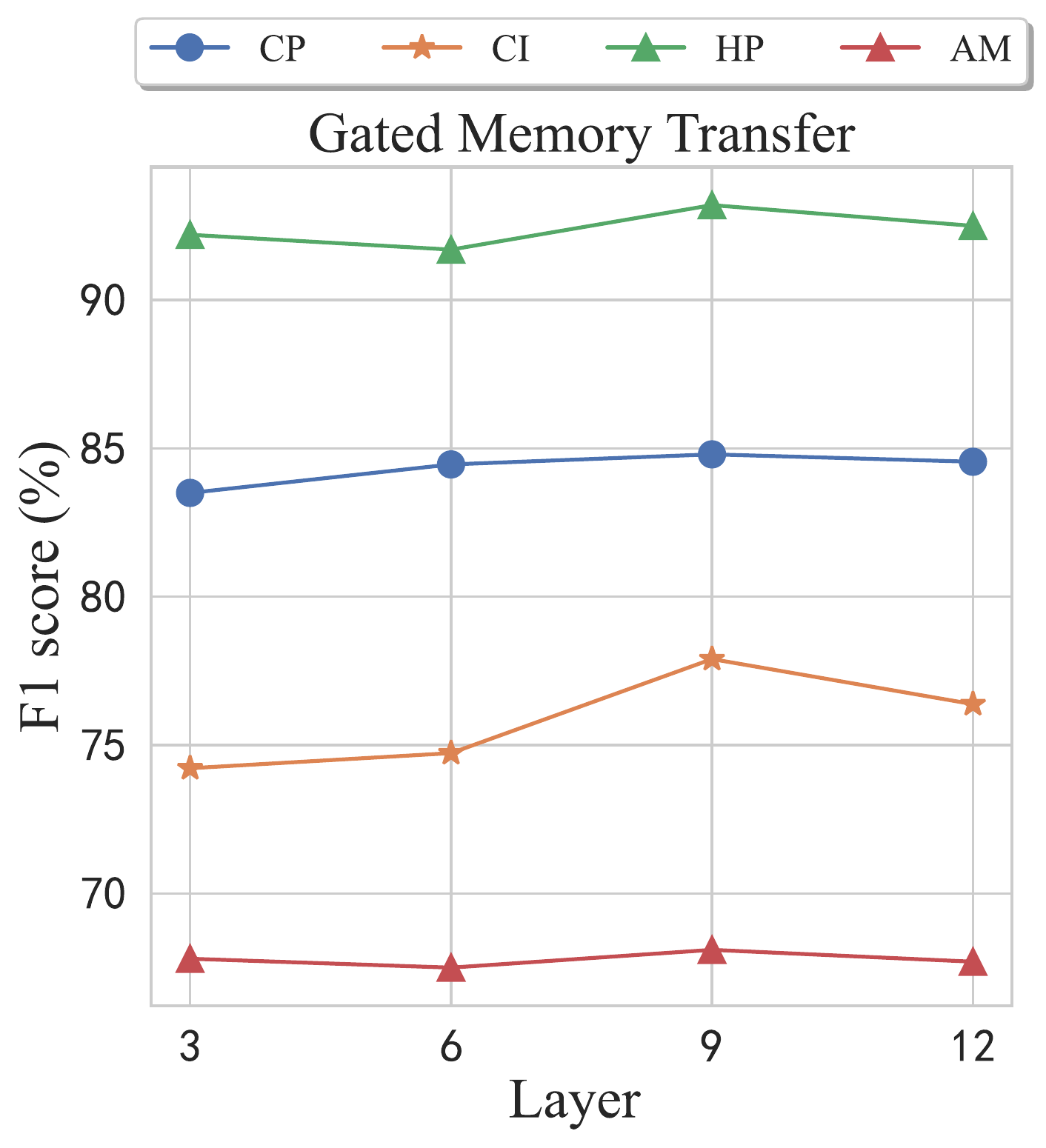}
\end{minipage}
\caption{Performance of different layer-selection indexes of memory-attention for single-layer memory transfer and gated memory transfer strategies.}\label{fig:single-gate}
\vspace{-0.2in}
\end{figure}

\section{Layer Selection Experiment}
\label{sec:layer_selection}
For single-layer memory transfer and gated memory transfer strategies, we experiment with adding the memory-attention to layers 3,6,9 and 12 in a 12-layer RoBERTa-base model, with the result shown in Figure~\ref{fig:single-gate}. We empirically find that adding memory-attention to the 9-th layer of the domain-specific model as the memory-augmented layer will obtain the best results for the two strategies. However, adding memory-attention to either too upper or too lower obtained fewer gains. Therefore, we adopt memory-attention on the 9-th layer as the default choice for the two strategies in the main experiment shown in Table~\ref{tab:main_table_1}.

\begin{table*}[h!]
\centering
\scalebox{0.75}{
\begin{tabular}{ccccccc}
\midrule[1.2pt]
\textbf{Task} & \textbf{Domain} & \textbf{Dataset} & \textbf{Train} & \textbf{Dev.} & \textbf{Test} & \textbf {Classes}\\
\hline 

\multirow{8}*{\textbf{Classification}}
  & BIOMED & \textsc{ChemProt}      & 4169 & 2427 & 3469 &  13 \\
  & BIOMED & \textsc{RCT}$^{\dag}$          & 180040 & 30,212 & 30135 & 5 \\
  & CS & \textsc{ACL-ARC}           & 1,688  & 114  & 139  & 6 \\
  & CS & \textsc{SCIERC}            & 3219 & 455 & 974 & 7 \\
  & NEWS &  \textsc{HyperPartisan}  & 515 & 65 & 65 & 2  \\
  & NEWS & \textsc{AGNews}$^{\dag}$           & 115000 & 5000  & 7600 & 4  \\
  & REVIEWS & \textsc{Amazon}$^{\dag}$         & 115251  & 5000 & 25000 & 2 \\
  & REVIEWS & \textsc{IMDB}$^{\dag}$           & 20,000 & 5000 & 25000 & 2 \\
    
\midrule
 \multirow{2}*{\textbf{Named Entity Recognition}} 
 
  & NEWS & CoNLL-2003      & 14,041  & 3,250 & 3,453 & -  \\
  & BIOMED & NCBI-Disease  & 5,433 & 924  & 941  & - \\

\midrule[1.2pt]
\end{tabular}
}
\caption{
Statistics of Classification and NER tasks involving four domains, including Biomedical, Computer Science, News, and Reviews. $^{\dag}$ indicates high-resource settings.
} \label{tab:statistic_data}
\end{table*}

%\scriptsize
\begin{table*} 
\centering
\scalebox{0.75}{
\begin{tabular}{lccccccc}
\midrule[1.2pt]
& & \multicolumn{2}{c}{ \textbf{Train} } & \multicolumn{2}{c}{ \textbf{Dev.} } & \multicolumn{2}{c}{ \textbf{Test} } \\
\textbf{Domain} & \textbf{Dataset} & \textbf{Context} &  \textbf{Question} &   \textbf{Context}  &\textbf{Question} &  \textbf{Context}  & \textbf{Question} \\

\midrule
REVIEWS & NewsQA & 11428  & 74160 & - & - & 106  & 674 \\
BIOMED & Medication & 182  & 7518 & 26  & 1858 & 53 & 4005 \\
\midrule[1.2pt]
\end{tabular}
}
\caption{
Statistics of QA tasks, including News and Biomedical domains. We report the number of contexts and questions of the two datasets. 
} \label{tab:statistic_data_2}
\end{table*}

\begin{table*} 
\centering
\scalebox{0.75}{
\begin{tabular}{lcccccccc}
\midrule[1.2pt]

&\multicolumn{2}{c}{ \textbf{BIOMED} } &\multicolumn{2}{c}{ \textbf{CS} } & \multicolumn{2}{c}{ \textbf{NEWS} } & \multicolumn{2}{c}{ \textbf{REVIEWS} }\\
\midrule
\textbf{Hyperparameters} & CP & RCT &  CI &   SE  & HP &  AG  & AM & IMDB\\

\midrule
Training Epochs & 14 & 4  & 15 & 15 & 12 & 5  & 6 & 15  \\
Batch Size per GPU     & 24 & 16  & 32 & 32 & 32 & 16  & 16 & 16  \\
Number of GPUs & 3 & 4  & 1 & 1 & 1 & 4  & 4 & 4  \\
Maximum Sequence Length & 256 & 256 & 256 & 256 & 256 & 384 & 512 & 512\\
Learning Rate & \multicolumn{8}{c}{ 4e-5 }  \\
Dropout  & \multicolumn{8}{c}{ 0.5 }  \\
Classification Layer & 1 & 1  & 1 & 1 & 1 & 2  & 1 & 2  \\
Learning Rate Optimizer &  \multicolumn{8}{c}{ Adam }  \\
Adam Epsilon & \multicolumn{8}{c}{ 1e-8 }  \\
Adam Beta 1  & \multicolumn{8}{c}{ 0.9}  \\
Adam Beta 2  & \multicolumn{8}{c}{ 0.999}  \\
\midrule[1.2pt]
\end{tabular}
}
\caption{Hyperparameters for fine-tuning on eight classification tasks of four domains, we use these hyperparameters for reporting the performances of our proposed G-MAP framework in the main papers.
} \label{tab:config_classfication}
\end{table*}

\begin{table*} 
\centering
\scalebox{0.75}{
\begin{tabular}{lcccc}
\midrule[1.2pt]
& \multicolumn{2}{c}{\textbf{BIOMED}} & \multicolumn{2}{c}{\textbf{NEWS}} \\
\textbf{Hyperparameters} & Medication & NCBI-Disease &  NewsQA&   CoNNL-2003 \\

\midrule
Training epochs & 3 & 20  & 3 & 15  \\
Batch size per GPU     & 16 & 32  & 16 & 32  \\
Number of GPUs  & 4 & 1  & 4 & 1  \\
Maximum Sequence Length & 384 & 128  & 384 & 128  \\
Classification layer & \multicolumn{4}{c}{ 1 }  \\
Learning rate optimizer & \multicolumn{4}{c}{ AdamW }  \\
Learning rate & 3e-5 & 5e-5  & 3e-5 & 5e-5  \\
Weight Decay & 1e-2 & 0  & 1e-2 & 0  \\
LR decay Warm-up rate & 6e-2 & 0  & 6e-2 & 0  \\
\midrule[1.2pt]
\end{tabular}
}
\caption{Hyperparameters for fine-tuning on QA and NER tasks of biomedical and news domains, we use them for reporting the performances of our proposed G-MAP framework in the main papers. 
} \label{tab:config_qa_ner}
\end{table*}

\end{document}